\documentclass[twoside,11pt]{article}

\usepackage[abbrvbib,preprint,nohyperref]{jmlr2e}

\usepackage{amssymb, cancel, amsmath, amstext}
\usepackage{epsfig}
\usepackage{natbib}
\usepackage{graphicx}

\usepackage{hyphenat}

\numberwithin{equation}{section}

\usepackage{mathtools}

\usepackage[all]{xy}
\usepackage[small]{caption}
\usepackage[american]{babel}
\usepackage[utf8]{inputenc}
\usepackage{url}
\usepackage{indentfirst}  
\usepackage{array}
\usepackage[normalem]{ulem}
\usepackage[margin=.9in, marginparwidth=.85in,marginparsep=.05in,heightrounded]{geometry}

\usepackage{tikz-cd}
\usepackage[scr=rsfso,cal=zapfc,frak=euler,bb=ams]{mathalfa}

\usepackage{mathrsfs}
\usepackage[shortcuts]{extdash}

\usepackage{booktabs}
\usepackage{caption}
\usepackage{subcaption}
\usepackage{colortbl}
\usepackage{mathrsfs}
\usepackage{multirow}
\usepackage{bbm}
\usepackage{bbold}
\usepackage[sans]{dsfont}
\usepackage[T1]{fontenc}
\usepackage{bm}

\usetikzlibrary{positioning}
\usepackage{stackengine} 
\newcommand\oast{\stackMath\mathbin{\stackinset{c}{0ex}{c}{0ex}{\ast}{\bigcirc}}}
\usepackage{tikz}

\usetikzlibrary{decorations.markings,plotmarks}
\usetikzlibrary{calc,trees,positioning,arrows,chains,shapes.geometric,%
decorations.pathreplacing,decorations.pathmorphing,shapes,%
matrix,shapes.symbols}
\tikzset{
>=stealth',
punktchain/.style={
rectangle, 
rounded corners, 
draw=black, very thick,
text width=25	em, 
minimum height=3em, 
text centered, 
on chain},
line/.style={draw, thick, <-},
element/.style={
tape,
top color=white,
bottom color=blue!50!black!60!,
minimum width=8em,
draw=blue!40!black!90, very thick,
text width=10em, 
minimum height=3.5em, 
text centered, 
on chain},
every join/.style={->, thick,shorten >=1pt},
decoration={brace},
tuborg/.style={decorate},
tubnode/.style={midway, right=2pt},
}
\tikzset{middlearrow/.style={
decoration={markings,
mark= at position 0.5 with {\arrow{#1}},
},
postaction={decorate}
}
}
\usepackage{makeidx}
\makeindex

\usepackage{color,soul}

\newcommand{\dt}{\Delta_{\mathrm{t}}}
\newcommand{\dtu}{\Delta_{\mathrm{t}}^{\mathrm{u}}}
\newcommand{\dtp}{\Delta_{\mathrm{t}}^{\mathrm{p}}}
\newcommand{\dx}{\Delta_{\mathrm{x}}}
\newcommand{\Nd}{\mathrm{N}_{\mathrm{d}}}
\newcommand{\Npt}{\mathrm{N}_{\mathrm{pt}}}

\newcommand{\Nu}{\mathrm{N}_{\mathrm{u}}}
\newcommand{\Np}{\mathrm{N}_{\mathrm{p}}}
\newcommand{\Nt}{\mathrm{N}_{\mathrm{t}}}
\newcommand{\Id}{\mathrm{Id}}
\newcommand{\Dd}{\mathrm{D}}

\newcommand{\li}{_{\ell^{\infty}}}

\newcommand{\R}{\mathbb{R}}

\newcommand{\N}{\mathbb{N}}
\newcommand{\Z}{\mathbb{Z}}
\newcommand{\ep}{\varepsilon}

\newcommand{\red}[1]{\textcolor{red}{#1}}

\usepackage[shortlabels]{enumitem}

\newtheorem{Lemma}{Lemma}

\newtheorem{Proposition}[Lemma]{Proposition}
\newtheorem{Corollary}[Lemma]{Corollary}
\newtheorem{Remark}[Lemma]{Remark}
\newtheorem{Problem}[Lemma]{Problem}
\newtheorem{Definition}[Lemma]{Definition}
\newenvironment{Proof}%
{\begin{trivlist} \item[]{\bf Proof. }}%
{\hspace*{\fill}$\rule{.4\baselineskip}{.4\baselineskip}$\end{trivlist}}

\DeclarePairedDelimiter{\fl}{\lfloor}{\rfloor}

\makeatletter
\def\U_#1{U^{\fl{#1}}\@ifnextchar[{\Ubrac}{\relax}}
\def\Ubrac[#1]{#1}
\makeatother

\makeatletter
\def\P_#1{P^{\fl{#1}}\@ifnextchar[{\Pbrac}{\relax}}
\def\Pbrac[#1]{#1}
\makeatother

\makeatletter
\def\W_#1{W^{\fl{#1}}\@ifnextchar[{\Wbrac}{\relax}}
\def\Wbrac[#1]{#1}
\makeatother

\makeatletter
\def\average_#1{\begin{minipage}[t]{1.5cm}
\centering
\emph{Average}\\
\small{$\partial\{#1\}$}
\end{minipage}}
\makeatother

\jmlrheading{1}{2021}{1-48}{4/00}{10/00}{meila00a}{Rafael Monteiro}

\ShortHeadings{Binary Classification as a Phase-Separation Process}{Rafael Monteiro}
\firstpageno{1}

\begin{document}
 \title{Binary Classification as a Phase Separation Process}

\author{\name Rafael Monteiro \email monteirodasilva-rafael@aist.go.jp,\\ 
\null \hfill                      \email rafael.a.monteiro.math@gmail.com \\
  \addr Mathematics for Advanced Materials Open Innovation Laboratory,\\
  AIST, c/o Advanced Institute for Materials Research,\\ 
  Tohoku University, Sendai, Japan}


\maketitle
\begin{abstract}%
We propose a new binary classification model called Phase Separation
Binary Classifier (PSBC). It consists of a discretization of a nonlinear reaction-diffusion equation coupled with an Ordinary Differential Equation, and is inspired by fluids behavior, namely, on how binary fluids
phase separate. Thus, parameters and hyperparameters have physical meaning, whose effects are studied in several different scenarios.

PSBC's equations can be seen as a dynamical system whose coefficients are trainable weights, with a similar architecture to that of a Recurrent Neural Network. As such, forward propagation amounts to an initial value problem. Boundary conditions are also present, bearing similarity with  figure padding techniques in Computer Vision.  Model compression is exploited in several ways, with weight sharing taking place both across and within layers.

The model is tested on pairs of digits of the classical MNIST database.  An associated multiclass classifier is also constructed using a combination of Ensemble Learning and one versus one techniques. It is also shown how the PSBC can be combined with other methods - like aggregation and PCA - in order to construct better binary classifiers. The role of boundary conditions and viscosity is thoroughly studied in the case of digits ``0'' and ``1''. 

\end{abstract}
\section{Introduction}\label{sec:introduction}
In practical terms, classification is a task that humans and machines perform in many different situations: deciding whether an article is worth reading or not, labeling an  image,   classifying a device as defective or functional, or, in an abstract fashion, assigning an object  $X$ to one of the $M$ classes  $\{ 0, \ldots, M -1\}$. When $M=2$ this process is called \textit{binary classification}, which will be the main focus of this paper.  

Several questions concerning classification are investigated in the field of Machine Learning (ML). Here we are interested in the particular case of binary classification using empirical risk minimization.  We refer to ``\cite{ProbabilisticPattern}'' for a more theoretical approach to supervised learning, pointing out additional references along the way.

In binary classification one assumes  the existence of an unknown map $h:\mathscr{X} \to \{0,1\}$, conveniently called \textit{hypothesis}, that one aims to investigate and somehow reconstruct, or approximate, from information available only on a subset of $\mathscr{X}$. In other words, given a data set (a sample) $\mathcal{D} := \{\left(X_{(i)}, Y_{(i)}\right)_{1\leq i \leq \Nd}\} \subset \mathscr{X}\times \{0,1\}$  and constraints
\begin{equation}\label{constraints}
h(X_{(i)}) = Y_{(i)}, 
\end{equation}
one wishes to construct a map $\widetilde{h}:\mathscr{X} \to \{0,1\}$ that is a good approximation to $h(\,\cdot\,)$ in a certain sense. It's quality can be measured for instance
\begin{equation}\label{accuracyformula}
 \text{Accuracy} =\frac{\text{cardinality}\left(\{ i \in \{1, \ldots, \Nd \}\,|\, Y_{(i)} = \widetilde{h}(X_{(i)})\}\right)}{\Nd},
\end{equation}
which must then be maximized.\,\footnote{Or minimizing the quantity $1 - \mbox{Accuracy}  = \frac{\text{cardinality}\left(\{ i \in \{1, \ldots, \Nd \}\,|\, Y_{(i)} \neq \widetilde{h}(X_{(i)})\}\right)}{\Nd}$, called  \textit{misclassification error}  \citep{ProbabilisticPattern}.}  

Since the constraints \eqref{constraints} are available for all $X_{(i)}$, this problem falls in the class of \textit{supervised learning}. Writing $h(\,\cdot\,) = \mathbbm{1}_{\mathcal{A}}(\,\cdot\,),$ where $\mathbbm{1}_{\mathcal{A}}(x) =1$ whenever $x \in \mathcal{A}$, $0$ otherwise. Hence, constructing an approximation to $h(\,\cdot\,)$ is equivalent to finding - or rather ``learning'' - the unknown set $\mathcal{A} = h^{-1}(\{1\})$. Once an approximation $\widetilde{h}: \mathscr{X} \to \{0,1\}$ is constructed, it can be applied to any element in $\mathscr{X}$ in order to ``predict'' whether it does, or does not, belong to $\mathcal{A}$; consequently, one refers to  $\widetilde{h}(\,\cdot\,)$  as a \textit{predictor}. Whenever a probability measure $\mu(\,\cdot\,)$ is considered in the space $\mathscr{X}\times \{0,1\}$, discovering $\mathcal{A}$ is called PAC learning;  \citep[Chapter 12]{ProbabilisticPattern}.

In the form just described the problem is too complex, for the  space of functions $\{g(\,\cdot\,)\, \vert\, g: \mathscr{X} \to \{0,1\}\}$ is too big and lacks mathematical structure. In practice, one follows the heuristics of choosing a smaller space of functions $ \mathscr{H}\subset \left\{g(\,\cdot\,)| g:\mathscr{X}\to \{0,1\}\right\}$ where $\widetilde{h}(\,\cdot\,)$ is sought for.  We call $\mathscr{H}$ a \textit{hypothesis space} \citep[\S 3]{cucker2002mathematical}.

\begin{figure}[htbp]
\centering
\begin{tikzcd}
Z^{\fl{0}} \arrow{r} & Z^{\fl{1}} \arrow{r}& Z^{\fl{2}} \arrow{r} & \hdots \arrow{r} & Z^{\fl{j}} 
\arrow{r} & \hdots \arrow{r}&Z^{\fl{Nt-1}}\arrow{r}& Z^{\fl{\Nt}}\\
W^{\fl{0}} \arrow{ur}& W^{\fl{1}}\arrow{ur} &W^{\fl{2}}\arrow{ur} &\hdots\arrow{ur}
& W^{\fl{j}}\arrow{ur}&\hdots \arrow{ur}&W^{\fl{\Nt -1}}\arrow{ur}&{}
\end{tikzcd}
\caption{Unfolded graph of a forward propagation in a network with $\Nt$ layers and trainable weights $W^{\fl{\cdot}}$. An arrow from $A$ to $B$ indicates that $B$ is a function of $A$.  \label{fig:fwdprop}}
\end{figure}
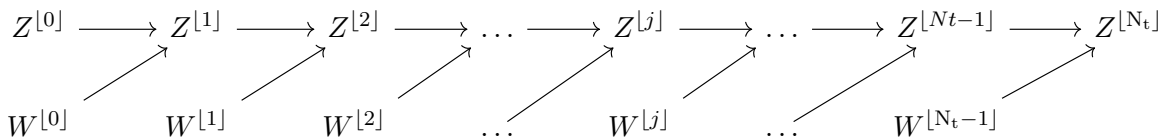

We briefly describe one of the techniques developed to construct the map $\widetilde{h}(\,\cdot\,)$, using  \textit{feedforward networks} \citep[Chapter 6]{DeepLearning}: these are graph structures devoid of cyclic loops, as seen in the diagram in Figure \ref{fig:fwdprop}.

For each $n \in \{0, \ldots, \Nt\}$ there exists an associated pair $\left(Z^{\fl{n}}, W^{\fl{n}}\right) \in \R^{z_{n}}\times\R^{w_n}$ referred to as a layer. When $n=0$ and $n= \Nt$, $Z^{\fl{0}}$ and $Z^{\fl{\Nt}}$ are called respectively input layer (that receives input data from $\mathscr{X}$) and output layer; when $1 \leq n \leq \Nt-1$, the vectors $Z^{\fl{n}}$ are called \textit{hidden layers}.  The variables $W^{\fl{0}}, \ldots, W^{\fl{\Nt -1}}$ are  referred to as \textit{trainable weights}, and are used for optimization purposes. Layers are all connected to each other in a hierarchical (that is, in  a tree-like) fashion as
\begin{equation}\label{fwdgeneral}
 Z^{\fl{n+1}} = \sigma^{\fl{n}}\left(Z^{\fl{n}}, W^{\fl{n}} \right), \quad \text{for }\quad 0 \leq n \leq \Nt-1.
\end{equation}
The maps $\sigma^{\fl{n}}:\R^{z_{n}}\times\R^{w_n} \to \R^{z_{n+1}}$ are called \textit{activation functions} and can be endowed with different properties as differentiability, decay in the far-field, etc \citep{HTFElements}. Last, one associates to this network a loss function $\mathscr{L}(Z, W, Y)$ that evaluates the sequence $Z=\left(Z^{\fl{0}}, \ldots, Z^{\fl{\Nt}} \right)$ and its proximity to a given label $Y \in \{0,1\}$ at a specific parameter value $W = \left(W^{\fl{0}}, \ldots, W^{\fl{\Nt -1}}\right)$. Using  the whole data set $\mathscr{D}$, these quantities make up a cost function $\mathrm{Cost}_{\mathscr{D}}(\,\cdot\,)$ of the form
\begin{equation*}
\mathrm{Cost}_{\mathscr{D}}(W) = \sum_{i=1}^{\Nd}\frac{\mathscr{L}\left(Z_{(i)}, W, Y_{(i)}\right)}{\Nd},
\end{equation*}
that one minimizes by optimization on $W$, as we explain next in detail.

Construction of $\widetilde{h}(\,\cdot\,)$ from data is called \textit{training} or \textit{model fitting}, and is  carried out using the network \eqref{fwdgeneral}. First, one generates an initial sequence of trainable weights $W_0 := \left(W^{\fl{0}}, \ldots, W^{\fl{\Nt -1}}\right)_0$ either randomly or deterministically. Then, for each pair of elements in $\mathscr{D} = \left\{(X_{(i)},Y_{(i)})_{1\leq i \leq \Nd}\right\} \in \mathscr{X}\times \{0,1\} = \R^{k_0}\times \{0,1\}$ one sets $Z^{\fl{0}} = X_{(i)}$ and generates a sequence $Z_{(i)}:=\left(Z^{\fl{0}}, \ldots, Z^{\fl{\Nt}} \right)_{(i)}$ using the feedforward network \eqref{fwdgeneral}. Afterwards, $W$ is updated using Gradient Descent, 
\begin{equation}\label{epochiteration}
W_{q+1} := W_{q} - \eta_q \frac{\partial \mathrm{Cost}_{\mathscr{D}}(W) }{\partial W}\Big|_{W = W_{q}},\quad q \in \N,
\end{equation}
where $W_q$ denotes the $q$-th iteration of this process.

\begin{figure}[htbp]
\centering
\begin{tikzcd}
 \frac{\partial \mathrm{Cost}}{\partial Z^{\fl{\Nt}}} \arrow{dr} \arrow{r} & \frac{\partial Z^{\fl{\Nt}}}{\partial Z^{\fl{Nt-1}}} \arrow{dr}\arrow{r} & \hdots \arrow{dr} \arrow{r} & \frac{\partial Z^{\fl{j}}}{\partial Z^{\fl{j-1}}} 
 \arrow{dr} \arrow{r} & \hdots \arrow{dr} \arrow{r}&\frac{\partial Z^{\fl{1}}}{\partial Z^{0}}\\
 {}& \arrow[dashed,red]{d}\frac{\partial Z^{\fl{\Nt}}}{\partial W^{\fl{Nt-1} }} &\hdots
 & \arrow[dashed,red]{d}\frac{\partial Z^{\fl{j}}}{\partial W^{\fl{j-1}}}&\hdots &\arrow[dashed,red]{d}\frac{\partial Z^{\fl{1}}}{\partial W^{\fl{0}}}\\
{}& \frac{\partial \mathrm{Cost}}{\partial W^{\fl{Nt-1} } } &\hdots
 & \frac{\partial \mathrm{Cost}}{\partial W^{\fl{j-1}}}&\hdots &\frac{\partial \mathrm{Cost}}{\partial W^{\fl{0}}}
\end{tikzcd}
\caption{Classical Backpropagation algorithm. Its development was a major improvement in the design of the algorithms to perform gradient descent. Full arrows for $A$  to $B$ denote composition ($A\circ B$). Dashed arrows from $A$ to $B$ denote variable assignment ($A \leftarrow B$); see further details in the Supplementary Material. \label{fig:backpropb}}  
\end{figure}
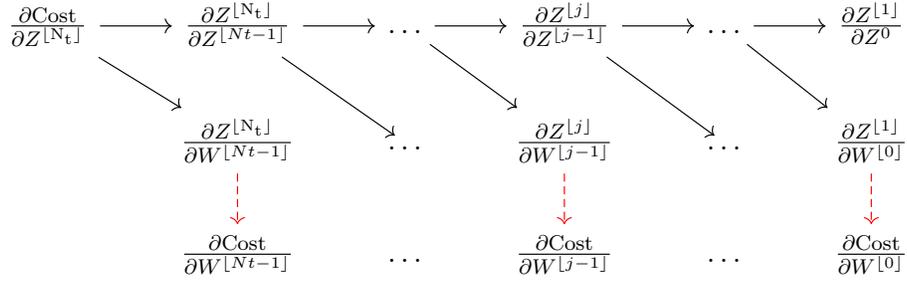

Each iteration \eqref{epochiteration} to update $W$  is called an \textit{epoch}. The quantities $\eta_{q}$ are positive numbers that receive the name of \textit{learning rates} and may vary across epochs. The computation of sequences $Z_{(i)}$ is called \textit{forward propagation}, while computing the derivatives of the cost with respect to $Z$ and $W$ is referred to as \textit{backpropagation}. Altogether, the algorithmic organization of this process receives the name of \textit{Backpropagation Algorithm} (BP), and is one of the cornerstones in the field of ML \citep[Chapter 6.5]{DeepLearning}; see Figure \ref{fig:backpropb}.

Once iterations satisfy  a stopping criteria (like tolerance, or some threshold), a parameter $W_*$ is obtained, and a class assignment, or \textit{discrimination rule}, takes place, corresponding in our case to
\begin{equation}\label{classassignment}
\begin{split}
\widetilde{h}\left(X_{(i)}\right) := \left\{\begin{array}{lll}
	  1, & \text{if} & \mathscr{L}(Z_{(i)},\,W_*,\,1) \leq \mathscr{L}(Z_{(i)},\,W_*,\,0),\\ 
	  0, & {} & \mathrm{otherwise},
	  \end{array}\right.
\end{split}
\end{equation}
finally completing the construction of $\widetilde{h}(\,\cdot\,)$. There are several challenges presented along the way though: the structure of the cost function's landscape can be very complex and in many applications optimizing with respect to $W$ consists of a non-convex optimization problem,  a difficulty that has received extensive attention from researchers in  recent years. 

Feedforward network architectures are ubiquitous in ML, appearing in Artificial Neural Networks (ANNs), Convolutional Neural Networks (CNNs), among other models, some of which  are considered analogies to the brain or human vision functioning. To what extent these models just mimic or are a faithful description of any biological processes is unclear and sometimes disputed \citep{MumfordIssues};  in this paper, however, we shall leave these considerations aside. We propose a new ML model whose motivation stems from a physical phenomenon by no means related to Biology nor (at least in principle) ``brain-like'' structures, but fluids and their dynamics: our model is inspired by how binary fluids phase separate. 

In full generality, the model propagates features using two variables that evolve simultaneously,
\begin{subequations}\label{fullmodel}
\begin{align}
 \frac{U_{m}^{\fl{n + 1}} - U_{m}^{\fl{n}}}{\dtu} &:= \frac{\ep^2}{\dx^2}\left(
U_{m+ 1}^{\fl{n + 1}} - 2 U_{m}^{\fl{n + 1}} + U_{m- 1}^{\fl{n + 1}}
\right) + f(U_m^{\fl{n}};\alpha_m^{\fl{n}} ), \quad \text{for } \quad 1\leq m \leq \Nu,\label{fullmodela}\\
\frac{ P_j^{\fl{n+1}} - P_j^{\fl{n}}}{\dtp} & := f(P_j^{\fl{n}}; \beta_j^{\fl{n}}), \quad \text{for } \quad 1\leq j \leq \Np,\label{fullmodelb}
\end{align}
\end{subequations}
where the nonlinearity  $f(u, w) = u(1 - u) (u-w)$ is of the same type in both systems. Parameters $\alpha^{\fl{\cdot}}$ in $\beta^{\fl{\cdot}}$ are ``trainable weights'', hence learned from data. We separate diffusion from non-linear effects by always taking $\dx^2 := \dtu$.  Initial conditions $\U_0$ are individuals' features, thus forward propagation amounts to an initial value problem, whereas $\P_0$ is a fixed vector. Note that diffusion  takes place in  feature space. 

\begin{Definition}[Phase Separation Binary Classifier - PSBC]\label{def:PSBC}
When a discrimination rule as in \eqref{classassignment} is allied the numerical scheme \eqref{fullmodel}, we  say that we have a  \textit{Phase Separation Binary Classifier}, denoted in short as PSBC. As we discuss below,   \eqref{fullmodel} falls in the class of feedforward networks, and in the sequel is also called PSBC model, whereas evolution of an initial condition is called forward propagation (see Figure \ref{fig:fwdprop}).  Whenever $\ep =0$  (resp., $\ep >0$) we refer to this classifier as  non-diffusive PSBC (resp., diffusive PSBC).  
\end{Definition}
It is  shown in Section \ref{sec:FullModel} that the feedforward networks' formulation \eqref{fwdgeneral} is sufficiently broad to encompass the PSBC model \eqref{fullmodel}, then written as
\begin{equation*}
\begin{split}
\left(\begin{array}{c}
\U_{n+1}\\
\P_{n+1}
\end{array}\right) := 
\sigma^{\fl{n}}\left(\U_n, W_{u}^{\fl{n}},\P_n, W_{p}^{\fl{n}}\right), \quad \text{with} \quad W_{u}^{\fl{n}} := \alpha^{\fl{n}}, \quad W_{p}^{\fl{n}} := \beta^{\fl{n}}. 
\end{split}
\end{equation*}
Nevertheless,  when compared with other state-of-art binary classifiers there are  striking differences, many of which we investigate and highlight throughout the text. For instance: the first term on the right hand side of \eqref{fullmodela}, a discretization of the Laplacian, consists of a diffusion operator acting on features; $f(\cdot,\cdot)$ is a nonlinear reaction term that is common in thresholding phenomena, and mostly plays a role on classification of an individual $X_{(i)}$ into one of the elements in  $\{0,1\}$. 

A quick description of each variable, without trying to fully elucidate their role at this point, goes as follows. The  scalars  $\dtu, \dtp>0$ and $\ep\geq0$  are \textit{hyperparameters} -  quantities that are neither optimized, nor inputs in $\mathscr{X}$ - and must be chosen in advance: the first two are due to time discretization, whereas the latter is a diffusion (or viscosity) term. We set $U^{\fl{0}} =X \in \mathscr{X}$, with $X$ representing features of an individual; thus, for each feature  $X_m$  we set $U_{m}^{\fl{0}}:= X_m$,  $1\leq m \leq \Nu$. Note that $U^{\fl{n}}\in \R^{\Nu}$ for all layers, where $\Nu$ is the Euclidean dimension of $\mathscr{X}$.  Boundary conditions of either Neumann or Periodic type affect the values of  $\U_{\cdot}_0$ and $\U_{\cdot}_{\Nu+1}$. A companion quantity $\P_{\cdot}$ evolves according to the ODE \eqref{fullmodelb} with initial condition $\P_0_j = \frac{1}{2}$ for $1 \leq j \leq \Np$; its role that is clarified in Section \ref{sec:coupling_in}. The vectors $\alpha^{\fl{\cdot}}\in \R^{\Nu}$ and $\beta^{\fl{\cdot}}\in \R^{\Np}$ contain variables that must be optimized according to a cost function. The number $\Np$ is either $1$ or a value $\Npt$ that indicates the  dimension of the subspace where  $\alpha^{\fl{\cdot}}\in \R^{\Nu}$ lies; see discussion in Section \ref{sec:reasoning}.

Throughout this paper  we  write $\U_{\cdot}(X;\alpha^{\fl{\cdot}})$ to represent the flow of $X$ through \eqref{fullmodela} or, in a different terminology, we say that $X$ is propagated by $\U_{\cdot}$; the same holds in the case $\P_{\cdot}(\frac{1}{2};\beta^{\fl{\cdot}})$ with respect to \eqref{fullmodelb}. In addition to that, we require a normalization condition of the initial data $X\in \mathscr{X}$, 
\begin{equation}\label{normalizationcondition}
 X \in [0,1]^{\Nu},
\end{equation}
easily obtained by data preprocessing. As we shall see in Section \ref{sec:1DsingleInvariantRegions}, the main reason to enforce \eqref{normalizationcondition} is  assuring that at any given epoch the forward propagation takes place inside an invariant region. 

In order to explain other qualities of the PSBC we need to understand the numerical scheme \eqref{fullmodel} and its contrasting differences  to other ML models. In the next section we present a succinct discussion of nonlinear diffusion equations, on which the mathematical and physical core of the PSBC model stand.

\subsection{Mathematical setting: nonlinear diffusion equations}
Clearly, the numerical scheme \eqref{fullmodela} is a semi-implicit finite-differences discretization of
\begin{equation}\label{AC}
\begin{split}
&\partial_t u(x,t) = \ep^2 \partial_{x}^2 u(x,t) + u(x,t) (1-u(x,t))(u(x,t) - \alpha(x)),\quad x\in [0,1],\\
&\partial_xu(0,t) = 0, \quad \partial_xu(1,t) = 0, \quad u(x,0) = u_0(x), 
\end{split}
\end{equation}
a nonlinear diffusion equation that describes phase separation in binary alloys and is known as \textit{Allen-Cahn equation}. Equation \eqref{AC} is one of the fundamental models in the theory of pattern formation \citep[IV.27]{ArndCompanion}. 

\begin{figure}[htb]

\begin{subfigure}[b]{.5\textwidth}
\centering  
\includegraphics[width=\textwidth]{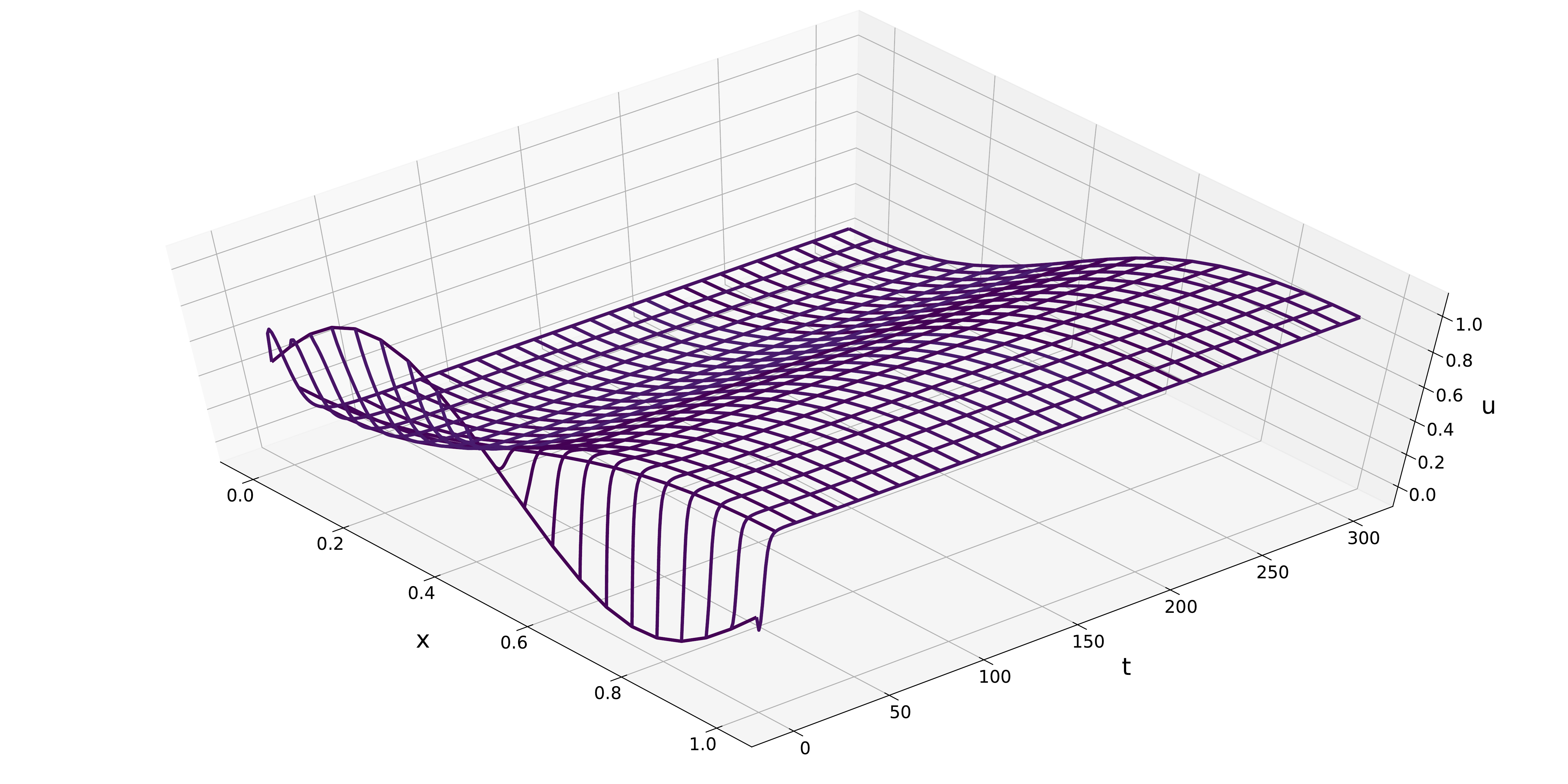}
\caption{$\alpha(x) = 4 - 8\left(x + 0.2\right)^2$.}  
\end{subfigure}
\hfill
\begin{subfigure}[b]{.5\textwidth}
\centering
\includegraphics[width=\textwidth]{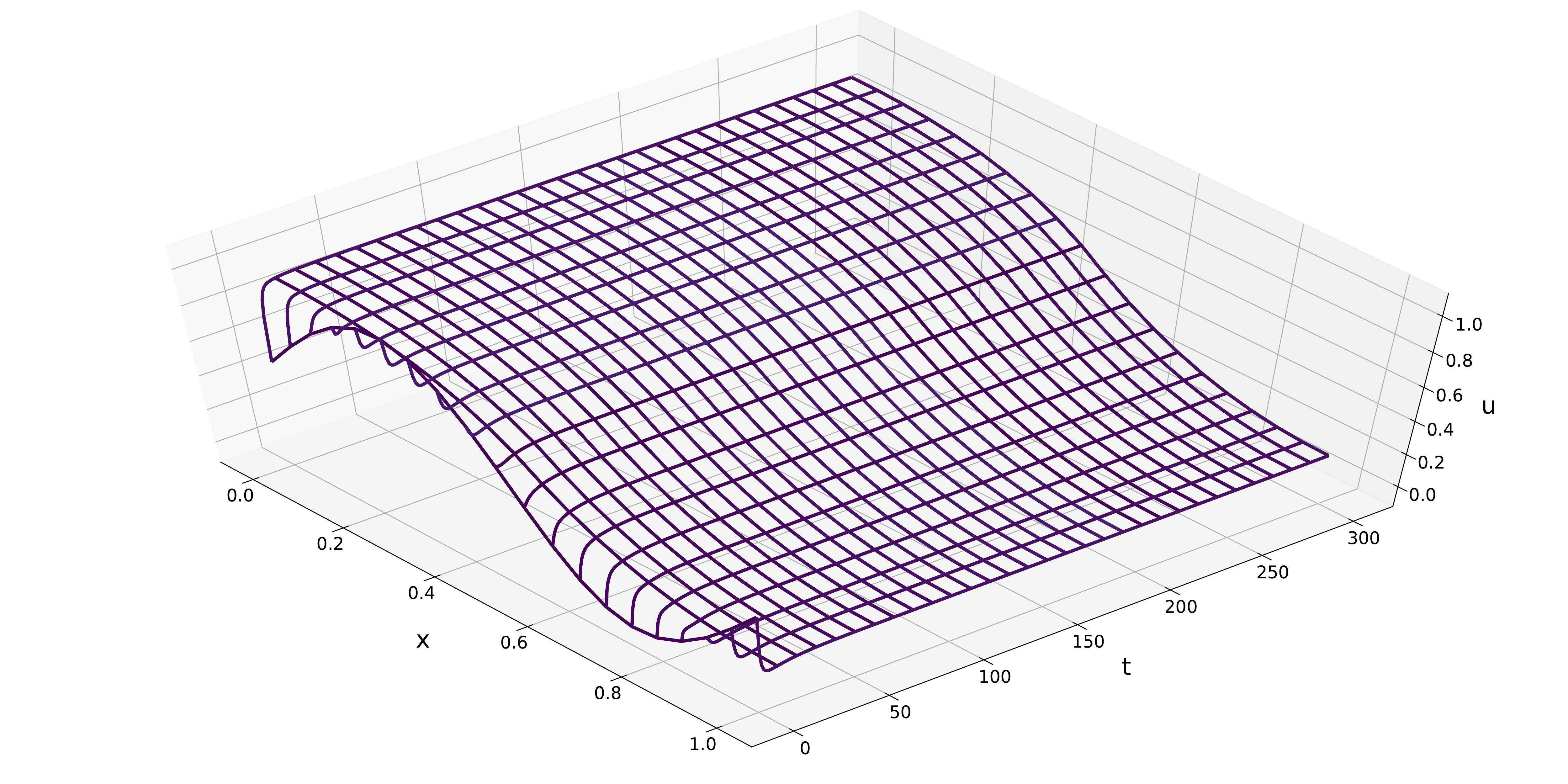}
\caption{$\alpha(x) = -2\,\cdot\left( \mathbbm{1}_{(-\infty,0.5)}(x)- \mathbbm{1}_{[0.5,+\infty)}(x)\right)$.}  
\end{subfigure}
\\
\begin{subfigure}[b]{.5\textwidth}
\centering
\includegraphics[width=\textwidth]{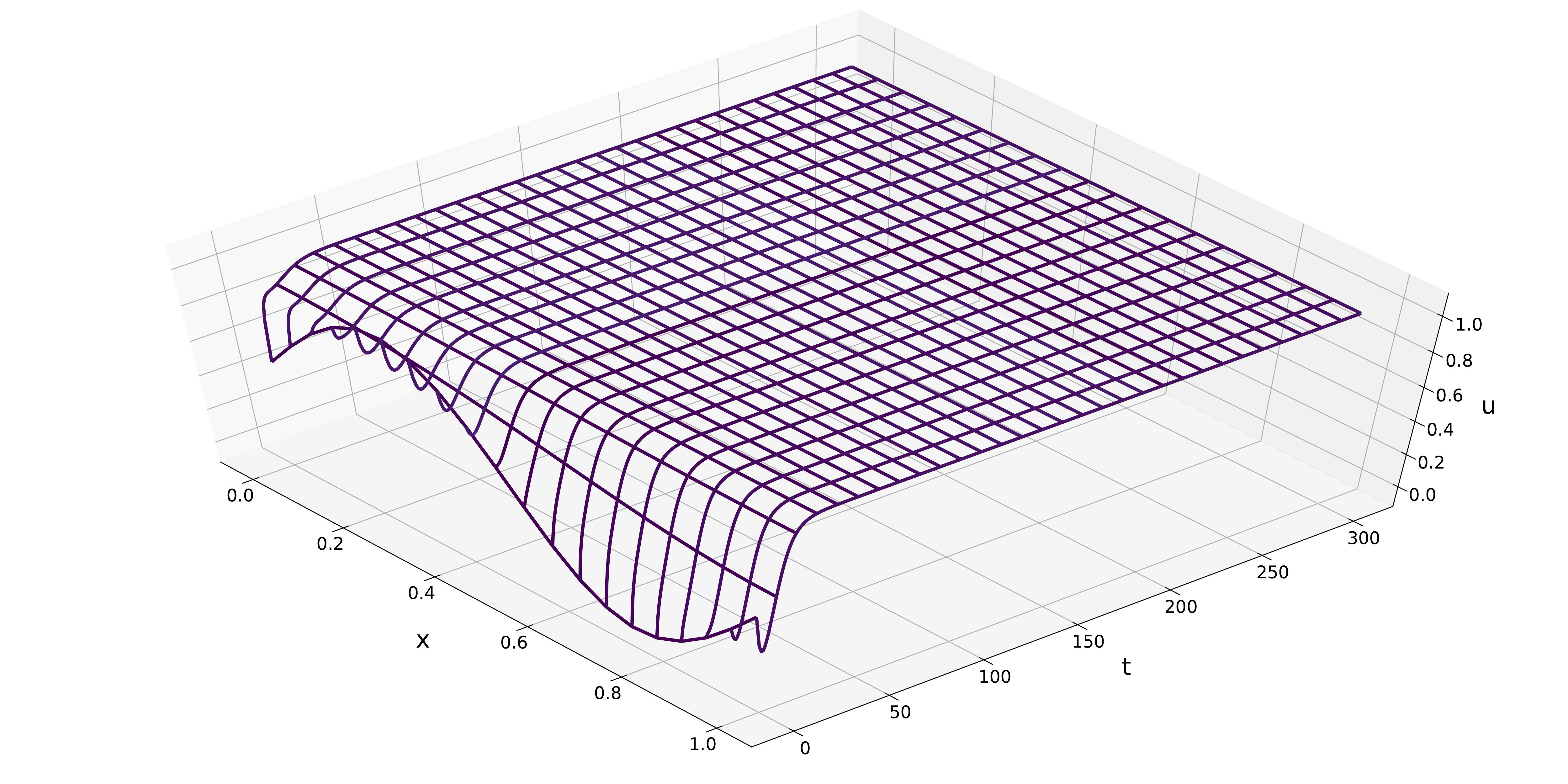}
\caption{$\alpha(x) = -0.8$.}   
\end{subfigure}
\hfill
\begin{subfigure}[b]{.5\textwidth}
\centering
\includegraphics[width=\textwidth]{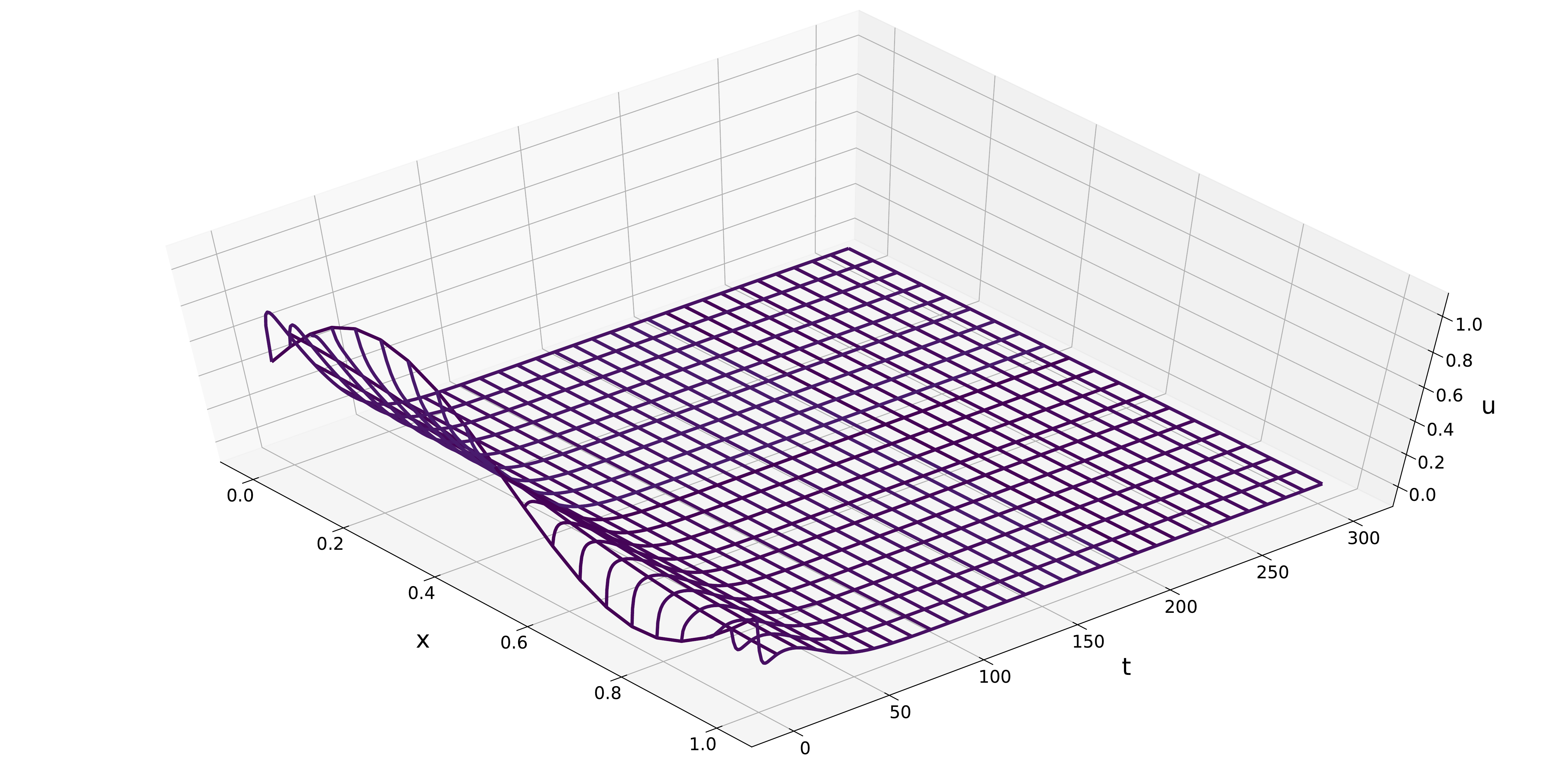}
\caption{$\alpha(x) = 0.9$.}  
\end{subfigure}
\caption{Numerical simulations of the Allen-Cahn Equation \eqref{AC} using  \eqref{fullmodela}  and different parameters $\alpha(\,\cdot\,)$. The initial conditions are fixed $u(x,t=0) = \frac{1}{2}-\frac{1}{2}\sin\left(\pi (2x -1)\right)$, on $x \in [0,1]$ with homogeneous Neumann boundary conditions. We use an uniform spatial grid with $\Nu=20$ points; $\Nt=300$, $\dtu =0.1$, and $\ep=0.3$. \label{fig:Neumann}}  
\end{figure}

In Equation \eqref{AC} time is represented by $t$, while $x$ represents space. The scalar $u(x,t)$ denotes a macroscopic quantity  that measures the relative proportion of two different species at $(x,t)$. The quantity $\ep \geq 0$ represents a diffusive term, and for this reason we call the model diffusive (resp., non-diffusive) whenever $\ep >0$ (resp. $\ep=0$). Finally, $\alpha(\,\cdot\,) \in \R$ describes the medium's spatial heterogeneity, being paramount to understanding how the initial value problem associated to  \eqref{AC} evolves;  some illustrative examples are shown in Figure \ref{fig:Neumann}. 

Following ``\cite{AngMal}'', one can  better understand the Allen-Cahn equation by first taking $\ep=0$ and $\alpha(\,\cdot\,)\in (0,1)$: in this case, for each $x \in [0,1]$ the dynamics in  \eqref{AC} decouples, yielding an ODE with two stable attracting points: $u \equiv 0$ and $u\equiv 1$. These limits are attained according to the initial state; namely, assuming that $\{x\in [0,1]\, |\, u(x,0) = \alpha(x)\}$ has measure zero, we have  convergence (almost everywhere) to
\begin{equation}\label{endstate}
\begin{split}
\lim_{t\to \infty}u(x,t) = \left\{\begin{array}{cc}
      1, & \quad \text{when} \quad u(x,0)> \alpha(x),\\
      0, & \quad \text{when} \quad u(x,0)< \alpha(x).
      \end{array}\right.
\end{split}
\end{equation}
Interestingly, the limiting function $\displaystyle{\lim_{t\to \infty}u(x,t)}$ assumes (almost everywhere) only two values, $0$ and $1$, in sharp contrast with $\alpha(\,\cdot\,)$ and the initial condition $u(\cdot,0)$, both $x$-dependent. If we bear in mind the previous discussion, one can imagine an initial condition $u(\cdot,0):= v^0(\,\cdot\,)$ as features of an individual with a class tag $Y\in \{0,1\}$, where $\alpha(\,\cdot\,)$ is a parameter measuring correlations among features;  as time evolves, the quantity $\alpha(\,\cdot\,)$ - in synergy with the nonlinearity $f(\cdot)$ - acts to classify $v^0(\,\cdot\,)$ in the ``correct'' way, which amounts to satisfying  $\displaystyle{\lim_{t\to \infty}u(\cdot, t;v^0) = Y}$. 

The mathematical study of  \eqref{AC} and other nonlinear diffusion equations is extensive. Under the framework of gradient-flows, it was shown in ``\cite{Chafee}'' that, whenever $\alpha(\,\cdot\,)$ is a constant $\alpha$, the only possible limit of  \eqref{AC} are constant solutions; consequently, in the case $0<\alpha <1$ this result implies that the only stable solutions are $u(\,\cdot\,) \equiv 0$ and $u(\,\cdot\,) \equiv 1$.  Later on the interplay between geometry and dynamics was brought to the limelight  in the seminal paper ``\cite{CastenHolland}'', where it was proven that  all stable solutions are constants also in the (spatial) multidimensional case, whenever the domain is convex \citep[Chapter 2]{Nidiffusion}.
%

There is a drastic change in the behavior of  \eqref{AC} when $\alpha(\,\cdot\,)$ is allowed to be non-homogeneous in space. In such case a larger class of stable non-constant stationary solutions exist:  it is shown in ``\cite{AngMal}'' that for some (non-constant) $\alpha(\,\cdot\,) \in \mathcal{C}^1([-1,1];\R)$ it is possible to construct stable stationary solutions $u(\,\cdot\,)$ that display several layers separating regions where $u(\,\cdot\,)$ is  either close to $0$ or close to $1$ \citep[Chapter 4, Sec. 4.3.8]{Hale}.

The previous discussion shows that different $\alpha(\cdot)$'s in \eqref{AC} yield different asymptotic behaviors of solutions. Naturally,  especially if we have binary classification in mind,  the following question arises: given some fixed $T^*>0$ (possibly $T^* = +\infty$), functions $v^0(\,\cdot\,)$ and $\displaystyle{v^{T^*}(\,\cdot\,)}$, is it possible to find a function $\alpha(\,\cdot\,)$ and an associated solution $u(x,t)$ to  \eqref{AC} such that 

\begin{align*}
 u(x,t)\Big\vert_{t=0}= v^0(x) \quad \text{and} \quad \displaystyle{ \lim_{t\uparrow T^*}u(x,t)}=v^{T^*}(x)?
\end{align*}
It is worth noticing that  \eqref{AC} is local but the equations to recover $\alpha(\,\cdot\,)$ are not, a common feature of inverse problems. 
With slightly more generality, we pose the previous question as a variational problem:

\begin{Problem}[Non-homogeneous $\alpha$ problem --- continuum version]\label{problem:abstract}
Let $\mathscr{B}$ and $\mathscr{A}$ be Banach spaces, and assume that for any initial conditions in $ v^0 \in \mathscr{X} \subset \mathscr{B}$ and $\alpha = \alpha(\,\cdot\,)\in \mathscr{A}$ the evolution model \eqref{AC} exists and is represented by $u(x,t;v^0)$. Given a family $\left\{\left(v_{(i)}^0(\,\cdot\,),v_{(i)}^{T^*}(\,\cdot\,)\right)_{i \in \Gamma}\right\}\in  \mathscr{X} \times \mathscr{T}\subset \mathscr{B} \times\mathscr{B}$ (possibly uncountable), find the best $\alpha(\,\cdot\,) \in \mathscr{A}$, if attainable, that minimizes 
\begin{equation}\label{TargetContinuum}
\lim_{t\uparrow T^*} \Vert u(\cdot, t;v_{(i)}^0) - v_{(i)}^{T^*}(\,\cdot\,)\Vert_{\mathscr{B}}.
\end{equation}
\end{Problem}
Roughly speaking, the goal is that of reconstructing  the heterogeneity of the media encoded $\alpha(\,\cdot\,)$. In other words, we  aim to  ``learn $\alpha(\,\cdot\,)$ from data'', where data consists of pairs of initial conditions $u(\, \cdot\,,t)\Big\vert_{t=0}= v^0(\,\cdot\,) \in \mathscr{X}$ and target functions $v^{T^*}(\,\cdot\,)\in \mathscr{T}$.  

A closely related but discrete formulation of  Problem  \ref{problem:abstract} is the backbone of the PSBC model. 
\begin{Problem}[Non-homogeneous $\alpha$ problem --- discrete version]\label{problem:numeric} 
Let $\Nt \in \mathbb{N}$. For all $0 \leq n \leq \Nt$, denote by $\left(\U_{\cdot}(X;\alpha^{\fl{\cdot}}), \P_{\cdot}(\frac{1}{2}\bm{1}_{\Np};\beta^{\fl{\cdot}})\right)\in \R^{\Nu} \times \in \R^{\Np}$  the solution to the discretized PDE \eqref{fullmodel}, of which $\alpha^{\fl{\cdot}} \in \R^{\Nu}$ and $\beta^{\fl{\cdot}} \in \R^{\Np}$ are parameters.
Given a map $\mathscr{F}:\R^{\Nu} \times \R^{\Np}\to \R^{\Nu}$ and a data set $\mathscr{D} = \{\left(X_{(i)}, Y_{(i)}\right)_{1\leq i \leq  \Nd}\} \subset \mathscr{X}\times \{0,1\}\subset \R^{\Nu}\times \{0,1\}$, for all $0\leq n \leq \Nt-1$ find $\left(\alpha^{\fl{n}},\beta^{\fl{n}}\right) \in \R^{\Nu}\times \R^{\Np}$ minimizing 
\begin{equation}\label{TargetSpace}
\mathrm{Cost}_{\mathscr{D}}\left(\alpha^{\fl{\cdot}}, \beta^{\fl{\cdot}}\right) = 
\sum_{i=1}^{\Nd}\frac{1}{\Nd}\left\Vert \mathscr{F}\left(\U_{\Nt}(X_{(i)}, \alpha^{\fl{\cdot}}), \P_{\Nt}(\frac{1}{2}\bm{1}_{\Np}, \beta^{\fl{\cdot}}  )\right) - Y_{(i)}\bm{1}\right\Vert_{\ell^{2}(\R^{\Nu})}^2,
\end{equation}
with discriminant function 
\begin{equation}\label{TargetSpace:discriminant}
\begin{split}
\widetilde{h}\left(X_{(i)}\right) := \left\{\begin{array}{ll}
            1, & \text{if} \quad \left\Vert \mathscr{F}\left(\widetilde{u},\widetilde{p}\right) - \bm{1}\right\Vert_{\ell^{2}(\R^{\Nu})}^2 \leq \left\Vert \mathscr{F}\left(\widetilde{u}, \widetilde{p} )\right)\right\Vert_{\ell^{2}(\R^{\Nu})}^2 ,\\ 
            0, & \mathrm{otherwise},
            \end{array}\right.
\end{split}
\end{equation}
where $(\widetilde{u}, \widetilde{p})  = \left(\U_{\Nt}(X_{(i)}, \alpha^{\fl{\cdot}}), \P_{\Nt}(\frac{1}{2}\bm{1}_{\Np}, \beta^{\fl{\cdot}}  )\right)$.
\end{Problem}
Let's clarify the similarities and differences between both problems. To begin with, imagine  a simpler scenario where   $\mathscr{F}(u, p) = u$. In this case,  \eqref{TargetSpace} becomes a discrete counterpart to \eqref{TargetContinuum}. In reality, it will be shown that a more general map $\mathscr{F}(\cdot,\cdot)$, non-trivial and relying on the companion Equation \eqref{fullmodelb}, is better suited for the classification task we are studying.

We should also observe that, unlike Problem \ref{problem:abstract}, the target space in Problem \ref{problem:numeric} consists of a much simpler - binary - set that  only contains the vectors $\bm{0}$ and $\bm{1}$ (respectively, vectors with only $0$'s, or  $1$'s).\,\footnote{This target space is not a vector space, just a subset of  $\ell^2(\R^{\Nu})$ with induced topology.} Discretization adds a high degree of flexibility to the problem, allowing both $\alpha^{\fl{\cdot}}$ and $\beta^{\fl{\cdot}}$ to either vary or repeat over layers. In this way, finding suitable trainable weights that minimize \eqref{TargetSpace} is related to problems in optimal control.

\begin{Remark}[Equivalent discriminant functions] It is more convenient to replace  \eqref{TargetSpace} by
\begin{equation}\label{TargetSpaceV2}
 \widetilde{\mathrm{Cost}}_{\mathscr{D}}\left(\alpha^{\fl{\cdot}}, \beta^{\fl{\cdot}}\right) = \sum_{i=1}^{\Nd}\frac{1}{\Nd}\left\vert \mathrm{Mean}\left(\mathscr{F}\left(\widetilde{u},\widetilde{p}\right)\right) - Y_{(i)}\right\vert^2,
\end{equation}
and \eqref{TargetSpace:discriminant} by the discriminant function 
\begin{equation}\label{TargetSpace:discriminantV2}
\begin{split}
\widetilde{\widetilde{h}}\left(X_{(i)}\right) := \left\{\begin{array}{ll}
            1, & \text{if} \quad \left\vert \mathrm{Mean}\left(\mathscr{F}\left(\widetilde{u},\widetilde{p}\right)\right) - \bm{1}\right\vert \leq \left\vert \mathrm{Mean}\left(\mathscr{F}\left(\widetilde{u},\widetilde{p}\right)\right)\right\vert ,\\ 
            0, & \mathrm{otherwise}.
            \end{array}\right.
\end{split}
\end{equation}
where $(\widetilde{u}, \widetilde{p})  = \left(\U_{\Nt}(X_{(i)}, \alpha^{\fl{\cdot}}), \P_{\Nt}(\frac{1}{2}\bm{1}_{\Np}, \beta^{\fl{\cdot}}  )\right)$ and  $\mathrm{Mean}(v)$ denotes the average over the vector $v$'s entries. There are two reasons for doing this. First,  because $\widetilde{h}(\cdot)$ and $\widetilde{\widetilde{h}}(\cdot)$ are in fact the same discriminant function, as  a direct expansion and rearrangement of  both sides in the inequality \eqref{TargetSpace:discriminant} shows. Second, because \eqref{TargetSpaceV2} is computationally much cheaper than \eqref{TargetSpace}. For these reasons we shall adopt \eqref{TargetSpaceV2} over \eqref{TargetSpace} in the rest of this paper.  
\end{Remark}
\subsection{Outline of the paper} \label{sec:IntroOutline}
This paper has two main goals: to introduce the PSBC, and to elaborate some of its fundamental mathematical properties, mostly by analysis of the numerical scheme \eqref{fullmodel}. Since there are many hyperparameters and architectures to be accounted for we begin by taking the whole classifier apart, gradually reassembling it piece by piece, elucidating each parameter's functionality along the way. %

\begin{figure}[htbp]
\centering
\begin{tikzcd}
Z^{\fl{0}} \arrow{r} & Z^{\fl{1}} \arrow{r}& Z^{\fl{2}}\hdots \arrow{r}&Z^{\fl{K-1}} \arrow{r}& Z^{\fl{K}} \arrow{r} & Z^{\fl{K+1}} \arrow{r} &\hdots\\
W^{\fl{0}} \arrow{ur}& W^{\fl{0}}\arrow{ur}&\hdots\arrow{ur} & W^{\fl{0}}\arrow{ur} &W^{\fl{K}}\arrow{ur}
&W^{\fl{K}}\arrow{ur} &{}
\end{tikzcd}
\caption{Unfolded graph of a forward propagation in a  layers-$K$-shared architecture with  $\Nt$ layers.  An arrow from $A$ to $B$ indicates that $B$ is a function of $A$. Whenever layer-$\Nt$-shared architectures are considered, the model falls in the class of Recurrent Networks.
\label{fig:FwdpropRepeatNJump}}
\end{figure}
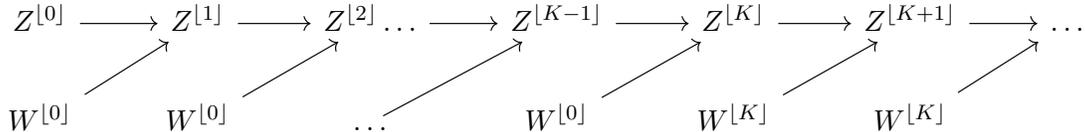

We begin in Section \ref{sec:reasoning}, explaining the reasoning behind the model. The PSBC exploits model compression through a wide range of weights sharing architectures \citep[Chapter 7.9]{DeepLearning}. 
\begin{Definition}[Weight sharing among layers]\label{def:LayersKShared}
We say that a feedforward  network  \eqref{fwdgeneral} with $\Nt$ layers  has \textit{layers-$K$-shared} property whenever
$$W = \left(W^{\fl{0}}, \ldots, W^{\fl{\Nt -1}}\right), \quad with \quad W^{\fl{i+j}} := W^{\fl{i}}\quad \text{for}\quad i \in K\Z,\quad j \in \{0,\ldots,K-1 \}.$$
\end{Definition}
Although the number of shared layers can take any value in the range $\{1, \ldots,  \Nt\}$, in this paper we concentrate  in two extreme cases: layers-$1$-shared (no shared layers; see Figure \ref{fig:fwdprop}), and layers-$\Nt$-shared (trainable weights repeat those in the first layer; see Figure \ref{fig:FwdpropRepeatNJump}).

Other types of weight sharing will also be exploited, but take place within layers. For instance, further reduction on the number of  trainable weights is attained by linear parameterization of $\left(\alpha^{\fl{\cdot}}\right)$  and   $\left(\beta^{\fl{\cdot}}\right),$  as
\begin{equation}\label{preview:BasisMatrix}
 \alpha^{\fl{k}} = \mathscr{B}_u W_u^{\fl{k}} , \quad \text{and}\quad \beta^{\fl{k}} = \mathscr{B}_p W_p^{\fl{k}},
\end{equation}
for $0\leq k \leq \Nt-1$, where $\mathscr{B}_u \in \R^{\Nu \times \Npt}$ and $\mathscr{B}_p \in \R^{\Nu \times \Np}$. %

Only a small set of parameters needs to be stored if both model's architecture and suitable implementation are simultaneously taken advantage of; in this fashion,  model compression enables efficient numerical parameterization. Weight sharing can also be seen as a regularization technique aiming to avoid over-fitting. Further motivation and technical details are given in Section \ref{sec:reasoning}
 
Still in Section \ref{sec:reasoning}, we turn our attention to numerical aspects of \eqref{fullmodel}. Since \eqref{fullmodela} is a semi-implicit (parabolic) scheme, there is an evident correspondence between forward propagation and  the initial value problem associated with \eqref{fullmodel}, therefore numerical stability issues have to be addressed.\,\footnote{ \label{theta_model} Indeed, the numerical scheme \eqref{fullmodela} can be seen as a particular case of
\begin{align*}
\U_{n+1} = \U_{n} +\theta \ep^2 \Dd_{\Nu}\U_{n} + (1 - \theta)\ep^2 \Dd_{\Nu}\U_{n+1} + \dtu\,f(\U_{n};\alpha^{\fl{n}} ). \quad \theta \in[0,1],
\end{align*}
If the nonlinearity is ignored, this model is known to be unconditionally stable when $0 \leq \theta \leq \frac{1}{2}$, that is, solutions remain stable for all values of $\dtu$ 
\citep[Section 6.3, page 147]{Strikwerda}. A similar numerical  scheme  has  also been exploited in  ``\cite{Hoff}''.} On one hand, the linear part of  \eqref{fullmodela} can be shown to be unconditionally stable for all values of $\dtu$ and $\dx$.  On the other hand, when nonlinearities are included, blow-ups of $U^{\fl{\cdot}}$ or $P^{\fl{\cdot}}$  are possible, and must be analyzed with care.
 
Forward propagation  also touches on the issue of global existence of the discrete dynamics \eqref{fullmodel}.  As trainable weights may vary from layer to layer, this question becomes even more challenging, paralleling that of a numerical scheme with variable coefficients. It is proven in Proposition \ref{prop:Globalpde} (and in earlier version of it, Proposition \ref{prop:GlobalODEsystem}) that one can control the $\ell^{\infty}$-norm of both $U^{\fl{\cdot}}$ and $P^{\fl{\cdot}}$ if constraints on $\dtu$ and $\dtp$ are imposed as
\begin{equation}\label{IREC}
0 < \dtu \leq \frac{1}{\sqrt{3}\, \mathrm{diameter}\left(\mathcal{R}_{\alpha}\right)^2}, \quad \mbox{and} \quad 0 < \dtp \leq \frac{1}{\sqrt{3}\, \mathrm{diameter}\left(\mathcal{R}_{\beta}\right)^2},
\end{equation}
where both $\mathcal{R}_{\alpha}$ and $\mathcal{R}_{\beta}$ are intervals constructed using $\left(\alpha^{\fl{k}}\right)_{0\leq k \leq \Nt-1}$,  and   $\left(\beta^{\fl{k}}\right)_{0\leq k \leq \Nt-1},$ respectively. For this reason we shall call the restrictions \eqref{IREC} on $\dtu$ and $\dtp$ \textit{Invariant Region Enforcing Conditions}.

Section \ref{sec:MNIST} is devoted to applications. We illustrate the PSBC's properties by applying it to the classical MNIST database, a benchmark data set commonly used to assess the quality of several different ML models \citep[]{Mnist}.  As a binary classifier, the model is evaluated  on pairs of digits. Binary classifiers are used to  construct a multiclass classifier in Section \ref{sec:multiclass}. A combination of Principal Components Analysis with the parameterization \eqref{preview:BasisMatrix} is discussed, indicating possible directions of improvement. An extensive study of the roles of  different boundary conditions,  parameterization cardinality $\Npt$, and diffusion term $\ep$ is carried out for the sub data set of digits ``$0$'' and  ``$1$''. 

\begin{figure}[htbp]
\centering
 \includegraphics[trim={0 8.5cm 0 8cm},clip,width=\textwidth]{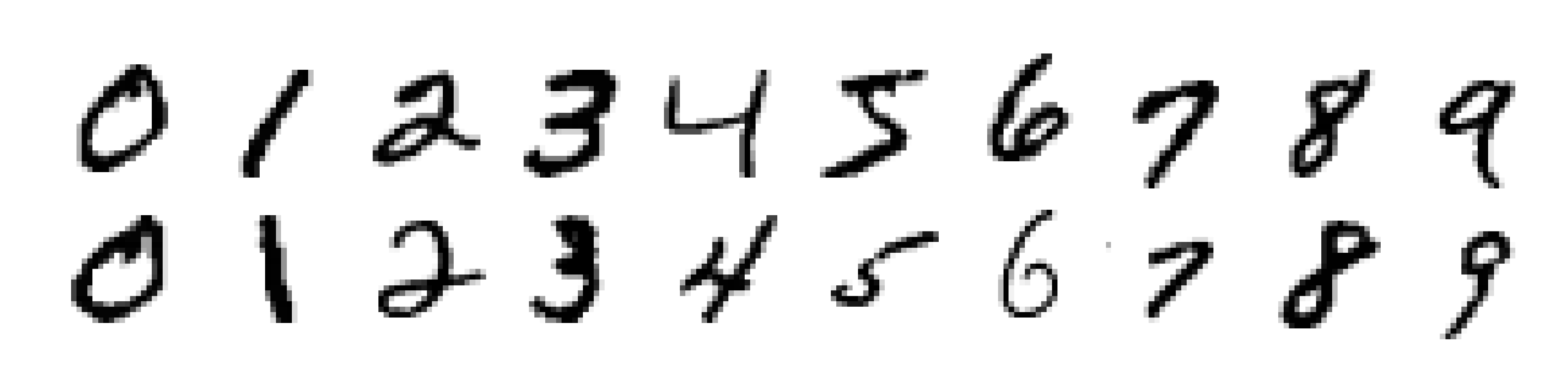}
 \caption{A sample from the MNIST database containing handwritten digits. Each picture has a shape $28 \times 28$ pixels, and is stored as a matrix; see the Appendix \ref{app:mnist} for further information. \label{fig2:Mnist}}
\end{figure}
 
An extensive discussion in Section \ref{sec:further_results} closes the paper with  several comments and open questions. Technical aspects of implementation and initialization of  hyperparameters are described in the Appendix \ref{app:initialization}.  Lengthier mathematical results are fully derived in the Appendix \ref{app:MP}, devoted to  discrete maximum principles, and in the Appendix \ref{app:proofs}. Additional figures are left to the Supplementary Material.\\

\subsection{Notation}\label{sec:notation}
Given a discrete Ordinary Differential Equation (ODE) or Partial Differential Equation (PDE), with initial condition $\U_{0}= X\in \R^{N}$ we denote its $n$-th iteration step by $\U_{n} = \U_{n}(X)= \left(U_m^{\fl{n}}\right)_{1\leq m \leq N}.$ 

Vectors and column matrices are identified: we write both $a \in \R^{n}$, as well as $a \in \R^{n\times 1}$. Given the Euclidean space $E = \R^{N}$, whenever $V \in E$ we say that $V \gtreqqless 0$ if $V_m\gtreqqless 0$ for all $1\leq m \leq N $. As such,  $V \in \R^{N}_{+}$  means $V\geq 0$. Averaging is written as $\mathrm{Mean}(V) = \frac{1}{N}\left(\sum_{m = 1}^{N}V_m\right).$
 
We write $\mathbb{G}_N:=\{1,\ldots, N\}$, calling  $\mathrm{supp}(V) = \{m \in  \mathbb{G}_N | V_m \neq 0 \}$ the support of $V$.

We  denote by $\bm{1} \in E$ (resp, $\bm{0} \in E$) a vector with all the entries $1$ (resp., $0$); whenever disambiguation is necessary, we use $\bm{1}_N$ or $\bm{0}_N$ to indicate the dimension of the underlying space these vectors are in. 

The canonical basis is defined as $e_i \in E$, $1 \leq i \leq N$. We shall further define
\begin{equation}\label{not:EA}
 e_{\mathcal{A}} := \sum_{i \in \mathcal{A}} e_i, \quad \text{for}\quad \mathcal{A} \subset \{1, \ldots, N\} = \mathbb{G}_N.
\end{equation}
Clearly, $e_{\{i\}} = e_i$, and $e_{\{1, \ldots, N\}}= \sum_{i=1}^{N} e_i = \bm{1} = \bm{1}_N$. By abuse of notation, we shall use \eqref{not:EA} for any Euclidean space, regardless of its dimension.

We shall say that a set $\mathscr{A}$ is convex if for any $p, q \in \mathrm{C}$ we have that $\lambda p + (1-\lambda) q \in \mathscr{A}$, for all $\lambda \in [0,1]$.   We write $\mathrm{conv}(S)$  to denote the convex hull of a set $S$, which corresponds to the intersection of all convex sets containing $S$; one can easily prove that the latter set is also convex. %

Last, ANNs, CNNs, and RNNs are acronyms for Artificial Neural Networks, Convolutional Neural Networks, and Recurrent Neural Networks, respectively. 
%
%
\section{The reasoning behind the PSBC  model}\label{sec:reasoning}
We begin our study of \eqref{fullmodel} by setting $\ep=0$, which yields an Euler discretization of an ODE,
\begin{subequations}\label{motivation}
 \begin{align}
 U_{m}^{\fl{n + 1}}  &:=  U_{m}^{\fl{n}} + \dtu  f(U_m^{\fl{n}};\alpha_m^{\fl{n}} ), \quad \text{for } \quad 1\leq m \leq \Nu,\quad \U_{0} = X \in \R^{\Nu}, \label{EqForu}\\
P_j^{\fl{n+1}} &:=   P_j^{\fl{n}} +\dtp f(P_j^{\fl{n}}; \beta_j^{\fl{n}}), \quad \text{for } \quad 1\leq j \leq \Np,\quad P^{\fl{0}} = \frac{1}{2} \bm{1}\in \R^{\Np}.\label{EqForp}
\end{align}
\end{subequations}
Since the nonlinearity  $f(u, w) = u(1 - u) (u-w)$ is the same in both models all the existence results discussed next apply to both equations. For this reason  we shall initially focus  on \eqref{EqForu}, postponing the discussion of the role of \eqref{EqForp} to Section \ref{sec:coupling_in}.

One of our main goals is verifying whether individual's features can be forward propagated through the network, namely, that $\U_{\cdot}(X,\alpha^{\fl{\cdot}})$ does not blow-up (causing a numerical overflow) before reaching the last layer of the network, where cost function evaluation and necessary optimization steps are taken. Notably and in contrast, models like  ANNs, CNNs, and RNNs bypass this question by using squashing (bounded) activation functions  (but not without trade-offs). In the PSCBC case, however, this concern  is legitimate because  Euler discretization methods do not have good stability properties \citep[Sec. 4.2]{Iserles}. Fortunately, we show in Proposition \ref{prop:GlobalODEsystem} that  control on the growth of $\Vert\U_{\cdot}(X,\alpha^{\fl{\cdot}})\Vert\li$ can be attained by adjustments on  $\dtu$ (under the Normalized Condition \ref{normalizationcondition}).\,\footnote{This is an example of nonlinear stabilization, since the region of A-stability of the Euler equations on the the region $\{z\in \mathbb{C}\,|\, \mathrm{Re}(z) \leq 0\}$ \citep[Chapter 4.2]{Iserles}.}  But first  we explain why it is possible to control the growth of solution in \eqref{motivation}, making a quick digression that takes us back to the  Allen-Cahn Equation \eqref{AC}.
\subsection{A glimpse on gradient flows and invariant regions on the continuum setting}\label{sec:1DsingleInvariantRegions}
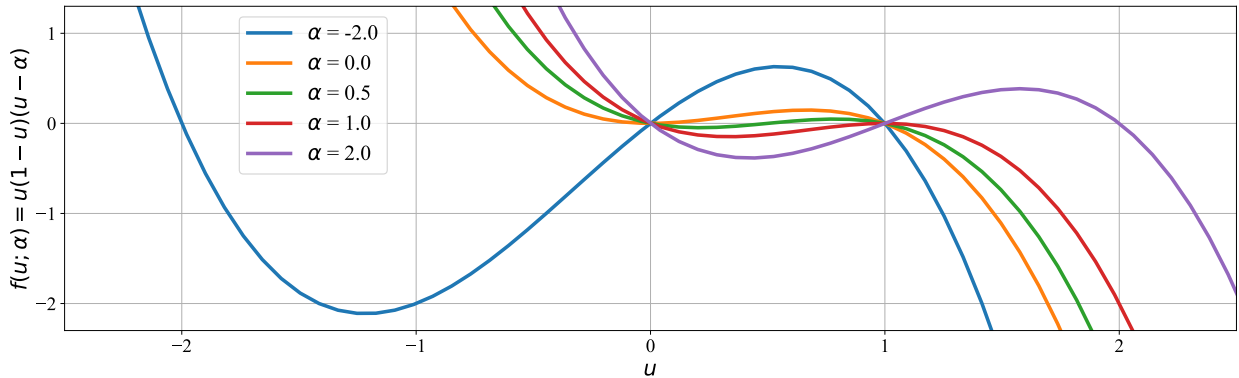
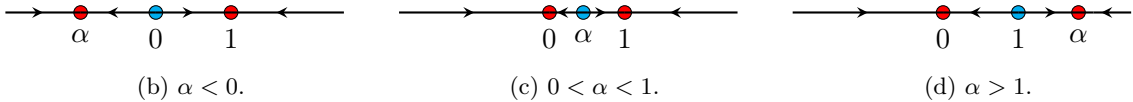
\begin{figure}[htb]
\centering
\begin{subfigure}[b]{\textwidth}
\centering
\includegraphics[width=\textwidth]{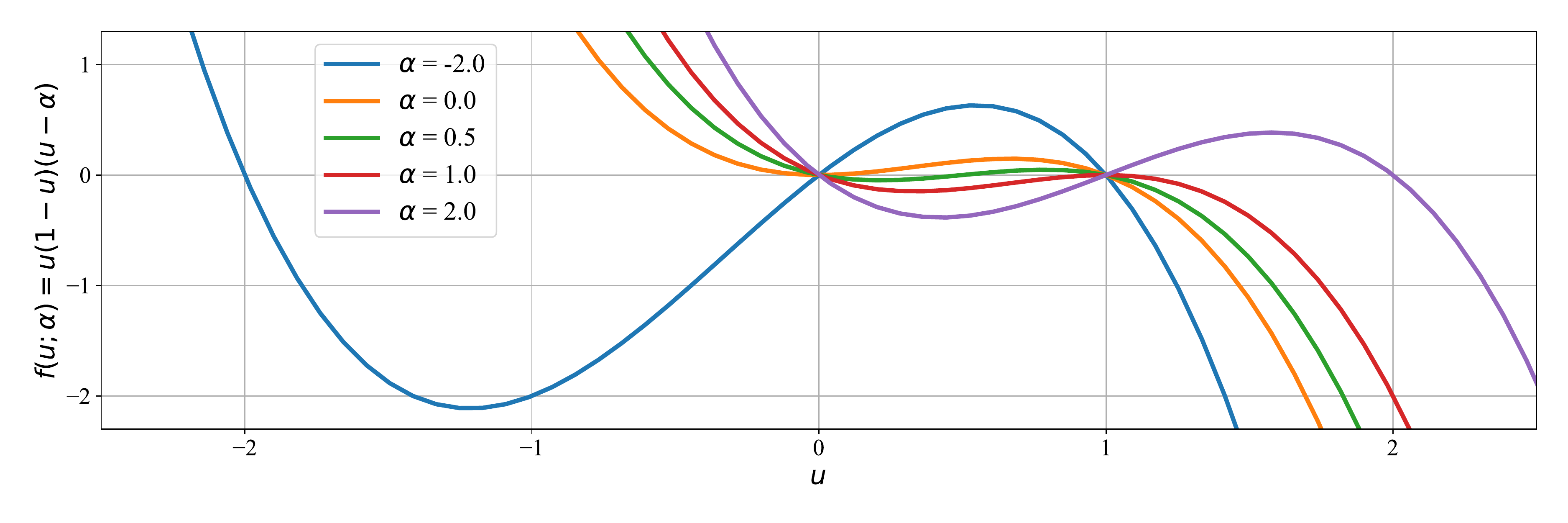}
\caption{Sketches of the function $\displaystyle{f(u;\alpha) =u(1-u)(u-\alpha)}$ for different values of $\alpha$. \label{fig:sketches_f}}
\end{subfigure}
\vspace{\lineskip}\\
\begin{subfigure}[b]{0.3\textwidth}
\begin{tikzpicture}[
middlearrow/.style 2 args={
decoration={    
  markings, 
  mark=at position 0.5 with {\arrow[xshift=3.333pt]{triangle 45}, \node[#1] {#2};}
},
postaction={decorate}
},
]
\tikzset{middlearrow/.style={
decoration={markings,
  mark= at position 0.5 with {\arrow{#1}},
},
postaction={decorate}
}
}

\draw node[draw=black, fill=red, circle,scale=.5, label=below:$\alpha$] at (-1,0) {};
\draw node[draw=black, fill=cyan, circle,scale=.5, label=below:0] at (0,0) {};
\draw node[draw=black, fill=red, circle,scale=.5, label=below:1] at (1,0) {};

\draw[thick,middlearrow={stealth}] (-2,0)--(-1,0);
\draw[thick,middlearrow={stealth reversed}] (-1,0)--(0,0);
\draw[thick,middlearrow={stealth}] (0,0)--(1,0);
\draw[thick,middlearrow={stealth reversed}] (1,0)--(2.5,0);
\end{tikzpicture}
\caption{$\alpha<0$. \label{fig:a_0}}
\end{subfigure}
\begin{subfigure}[b]{0.3\textwidth}
\begin{tikzpicture}[
middlearrow/.style 2 args={
decoration={    
  markings, 
  mark=at position 0.5 with {\arrow[xshift=3.333pt]{triangle 45}, \node[#1] {#2};}
},
postaction={decorate}
},
]
\tikzset{middlearrow/.style={
decoration={markings,
  mark= at position 0.5 with {\arrow{#1}},
},
postaction={decorate}
}
}

\draw node[draw=black, fill=red, circle,scale=.5, label=below:0] at (0,0) {};
\draw node[draw=black, fill=cyan, circle,scale=.5, label=below:$\alpha$] at (.45,0) {};
\draw node[draw=black, fill=red, circle,scale=.5, label=below:1] at (1,0) {}; 
\draw[thick,middlearrow={stealth}] (-2,0)--(0,0);
\draw[thick,middlearrow={stealth reversed}] (0,0)--(.5,0);
\draw[thick,middlearrow={stealth}] (.5,0)--(1,0);
\draw[thick,middlearrow={stealth reversed}] (1,0)--(2.5,0);
\end{tikzpicture}
\caption{$0<\alpha<1$. \label{fig:0a1}}
\end{subfigure}
\begin{subfigure}[b]{0.3\textwidth}

\begin{tikzpicture}[
middlearrow/.style 2 args={
decoration={    
  markings, 
  mark=at position 0.5 with {\arrow[xshift=3.333pt]{triangle 45}, \node[#1] {#2};}
},
postaction={decorate}
},
]
\tikzset{middlearrow/.style={
decoration={markings,
  mark= at position 0.5 with {\arrow{#1}},
},
postaction={decorate}
}
}
\draw node[draw=black, fill=red, circle,scale=.5, label=below:0] at (0,0) {};
\draw node[draw=black, fill=cyan, circle,scale=.5, label=below:1] at (1,0) {};
\draw node[draw=black, fill=red, circle,scale=.5, label=below:$\alpha$] at (1.8,0) {}; 
\draw[thick,middlearrow={stealth}] (-2,0)--(0,0);
\draw[thick,middlearrow={stealth reversed}] (0,0)--(1,0);
\draw[thick,middlearrow={stealth}] (1,0)--(2,0);
\draw[thick,middlearrow={stealth reversed}] (2,0)--(2.5,0);
\end{tikzpicture}
\caption{$\alpha>1$. \label{fig:1a}}
\end{subfigure}

\caption{Phase portraits of the ODE $\frac{d}{dt} u(t) = f(u;\alpha) = u (1-u)(u - \alpha)$, for different values of $\alpha$, with red (resp., blue) circles representing stable (resp., unstable) stationary points. Notice that in all the cases presented, the lowest and largest stationary points are stable, a fact that is exploited in the search for invariant regions, as explained in Section \ref{sec:coupling_in}. \label{fig:phase_portrait}}
\end{figure}

As said above, when $\Nu = 1$ the numerical scheme \eqref{motivation} is a discretization of the ODE
\begin{equation}\label{ODEModel1:continuum}
\frac{d u}{dt} = f(u; \alpha) := u(1-u)(u-\alpha), \quad u(X, t=0) = X \in \R,\quad \alpha \in \R.
\end{equation}
The nonlinearity in  \eqref{ODEModel1:continuum} is of bistable type, with 3 stationary points: $u=0$, $u=1$, and $u=\alpha$. When $\alpha$ varies, the qualitative behavior of the stationary points $0$ and $1$ change, as we can see on the phase portrait of the ODE \eqref{ODEModel1:continuum} in Figure \ref{fig:phase_portrait}. 

\noindent
One can associate an energy functional $\displaystyle{\mathscr{E}(u,\alpha) = -\int^u s(1-s)(s-\alpha)\mathrm{ds}}$ to \eqref{ODEModel1:continuum}, which now reads
\begin{equation*}
\frac{d u}{dt} = f(u; \alpha) := -\partial_u\mathscr{E}(u, \alpha), \quad\text{where} \quad u(X,t=0) = X \in \R,\quad \alpha \in \R.
\end{equation*}
Consequently, along trajectories of \eqref{ODEModel1:continuum} we must have 
\begin{equation}\label{dissipative}
\frac{\partial}{\partial t}\mathscr{E}(u, \alpha) = - \vert\nabla_u \mathscr{E}(u, \alpha) \vert^2 \leq 0,
\end{equation}
which can  be interpreted as the rate of energy dissipated as the system evolves; these ideas and concepts can be generalized to systems and other functional spaces, and are an important tool in characterization of rate of convergence to asymptotic states \citep[Sec. 2]{carrillo2006contractions}. 

The phase portraits in Figure \ref{fig:phase_portrait} indicate that  the lowest and the largest stationary points of \eqref{ODEModel1:continuum} are always either stable or semi-stable, regardless of the value $\alpha$. In fact, using \eqref{dissipative} and the structure of $\mathscr{E}(\cdot, \alpha)$, we can conclude that any interval $\mathscr{I}$ containing the set
\begin{equation}\label{SigmaAlpha}
 \Sigma_{\alpha} :=\left[\, \min\{0,\alpha\},\, \max\{1,\alpha\} \,\right]
\end{equation}
is positively invariant with respect to \eqref{ODEModel1:continuum}, namely, whenever $x_0 \in \mathscr{I} \supset \Sigma_{\alpha}$ and $\mathscr{I}$ is an interval, it holds that $u[x_0,t]\in \mathscr{I}$ for all $t\geq 0$ \citep[Chapter 9]{linear_algebra-smale-hirsch}. Similar  ideas have been applied in PDEs, as discussed in  Remark \ref{rmk:chueh_conway_smoller_hoff}.

Naturally, this whole discussion readily extends  to cases in which features lie in a high dimensional feature space. Indeed, with  $\mathscr{E}(\cdot; \cdot)$ as given in Section \ref{sec:1DsingleInvariantRegions} and $U \in \R^{\Nu}$ for any $\Nu \geq 1$, we can define a map $\widetilde{\mathscr{E}}: \R^{\Nu}\times \R^{\Nu}\to \R$
\begin{equation}\label{SystemGdFlow}
\frac{d U}{d t} = - \nabla_U \widetilde{\mathscr{E}}(U,\alpha), \quad \text{where} \quad \widetilde{\mathscr{E}}(U,\alpha) = \sum_{i = 1}^{\Nu} \mathscr{E}(U_i; \alpha_i),
\end{equation}
which describes the dynamics (of features) as a gradient flow.
\subsection{Back to the discrete setting: model compression and existence of dynamics $(\ep =0)$}
\paragraph{Weight sharing within layers I.} Although each entry of $\alpha^{\fl{\cdot}}$ in \eqref{EqForu}   can be optimized independently, a certain degree of flexibility - and model compression - can be achieved by parameterization of $\alpha^{\fl{\cdot}}$; in this way we simultaneously   generalize the model and reduce the  number of trainable weights. This can be  achieved by considering  
\begin{equation}\label{BasisMatrix}
\alpha^{\fl{n}} = \mathscr{B}_{u} W_{u}^{\fl{n}}, \quad \text{with} \quad  \mathscr{B}_{u} \in \R^{\Nu \times \Npt} \quad \text{and} \quad W_{u}^{\fl{n}} \in \R^{\Npt}.  
\end{equation}
In this paper we fix $\mathscr{B}_{u}$ over layers, although this restriction can be removed. We refer to $\Npt$ as  \textit{ parameterization cardinality} and to $\mathscr{B}_{u}$ as a \textit{basis matrix}. 

The specific form of $\mathscr{B}_{u}$ is important and may be chosen according to the prediction problem; other possibilities are contemplated in Section \ref{sec:pca}.
As we show next, under appropriate conditions on $\mathscr{B}_{u}$ the $\ell^{\infty}$ norm of  $\alpha^{\fl{\cdot}}$ and $W_{u}^{\fl{\cdot}}$  are proportional.

\begin{Lemma}
In order to avoid over-parameterization,  assume that $\Npt \leq \Nu$ and $\mathscr{B}_{u} = [b_1 | \ldots | b_{\Npt}]$ has full rank (i.e., its columns are linearly independent). Then:

\begin{enumerate}[label=(\roman*), ref=\theTheorem(\roman*)]
\item \label{BasisMatrix:param} (Parameterization proportionality)
It holds that $\Vert\alpha^{\fl{n}}\Vert\li\approx\Vert W_{u}^{\fl{n}}  \Vert\li $, with constants depending on $\mathscr{B}_{u}$. 
\item \label{BasisMatrix:decop} (Decoupled evolution in terms of $\mathscr{B}_{u}$) The  dynamics $\U_{\cdot}(X)$ of any initial condition $X$ through \eqref{EqForu} is so that each feature evolves independently. Furthermore, if  $m \in \mathbb{G}_{N}$  is such that  $m \not \in \mathrm{supp}(b_i)$, then the values in $b_i$ do not affect the evolution of $\U_{\cdot}_m(X)$.
\end{enumerate}
 
\end{Lemma}
The proof of (i) is immediate after rewriting \eqref{BasisMatrix} in its normal form  $w =\left(\mathscr{B}_{u}^{T}\mathscr{B}_{u}\right)^{-1}\mathscr{B}_{u}\alpha$. Inspection of \eqref{EqForu} and \eqref{BasisMatrix} yields (ii). Lemma \ref{BasisMatrix:param} shows that lower or upper bounds in terms of  either $\alpha^{\fl{\cdot}}$ or its parameters $W_u^{\fl{\cdot}}$ are equivalent. Lemma \ref{BasisMatrix:decop} on the other hand shows that the model has a parallelized architecture (see Remark \ref{rmk:parallelization}).

\begin{Remark}[A canonical basis matrix construction] \label{rem:BasisMatrix} Let $1\leq \Npt \leq \Nu$, in such a way that $\Nu = Q\Npt$, for $Q \in \N$. Initially, set $\pi_0 := \emptyset$. For $1\leq j \leq \Npt$, construct  sets $\pi_j$ recursively with the  $Q$ smallest elements in the set $\mathbb{G}_{\Nu} \setminus \left(\cup_{l = 0}^{j-1} \pi_l\right)$. Finally, define $\mathscr{B} := \begin{bmatrix} e_{\pi_1} |\ldots| e_{\pi_{\Npt}} \end{bmatrix}$ (recall notation in Section \ref{sec:notation}). For example,  when $\Npt = \Nu$ we get $\mathscr{B} = \Id_{\Nu}$, whereas when $\Npt = 1$ we obtain $\mathscr{B} = \bm{1}_{\Nu}$.
We stress that under this construction Lemma \ref{BasisMatrix:param} holds as an equality, namely,  $\Vert\alpha^{\fl{\cdot}}\Vert\li = \Vert W_{u}^{\fl{\cdot}}\Vert\li.$ Other cases (like $\Nu$ and $\Npt$ not multiple) are constructed similarly, and  given in full generality in the Supplementary Material.  
\end{Remark}

\paragraph{Forward propagation $(\ep = 0)$ --- existence of dynamics} We shall prove in this section,  under mild conditions on $\dtu$ and the models' weights,  that features can be forward propagated through the network and remain bounded. This is achieved  by proving the existence of a bounded box where features start at - and remain inside - as the dynamics through \eqref{EqForu} unfolds.

The existence of such an invariant region  immediately implies full control of lower and upper bounds of $u(t;\cdot)$ for all $t\geq 0$. Overall, it  gives some type of stability of solutions up to possibly  oscillatory behavior. Although the  gradient flow structure is mostly lost upon discretization, it is shown in the next proposition that it can still be taken advantage of. Interestingly,  numerical discretization yields an invariant set  that is slightly bigger than the interval $\Sigma_{\alpha}$ in \eqref{SigmaAlpha}. 

\begin{Proposition}[Global existence of forward propagation --- inviscid case ($\ep =0$)]\label{prop:GlobalODEsystem} Let $\Nt \in \N$ and  $\U_{0} :=X\in [0,1]^{\Nu}$.
Given two sequences of trainable weights
$\left(\alpha^{\fl{n}}\right)_{0\leq n \leq \Nt-1} \in \R^{\Nu}$, 
augmented by  $\alpha^{\fl{-1}} := \bm{1} \in \R^{\Nu}$, denote by  $\U_{\cdot}(X, \alpha^{\fl{\cdot}})$ the values  obtained using \eqref{prop:GlobalODEsystem}.
For any fixed $1\leq v \leq \Nu$ and $-1 \leq k \leq \Nt-1$, define the quantities
\begin{align*}
L_{\alpha,v}^{\fl{k}} := \inf_{-1\leq n \leq k}\left(\min\left\{\alpha_v^{\fl{n}},0\right\}\right),\quad R_{\alpha,v}^{\fl{k}} := \sup_{-1\leq n \leq k}\left(\max\left\{\alpha_v^{\fl{n}},1\right\}\right).
\end{align*}
Assume that $-\infty < L_{\alpha,v}^{\fl{\Nt-1}}$ and $R_{\alpha,v}^{\fl{\Nt-1}}<+\infty$ hold, and that the Invariant Region Enforcing Condition \eqref{IREC} applies as
\begin{equation}\label{prop:Globalode:dt}
0\leq \dtu \leq \min_{1\leq v \leq \Nu}\frac{1}{\sqrt{3}\left(\vert L_{\alpha,v}^{\fl{\Nt-1}}\vert + \vert R_{\alpha,v}^{\fl{\Nt-1}}\vert \right)^2}. 
\end{equation}
Then, for any $1\leq v \leq \Nu$ and $1\leq q \leq \Np$ the sequences $\U_k_v(X, \alpha^{\fl{k}})$ and $\P_k_q(X, \beta^{\fl{k}})$ satisfy
\begin{equation}\label{prop:GlobalODEsystemIneq}
L_{\alpha,v}^{\fl{k-1}} - \dtu M_{\alpha,v}^{\fl{k-1}} \leq \U_{k}_v(X, \alpha^{\fl{k}}) \leq R_{\alpha,v}^{\fl{k-1}} + \dtu M_{\alpha,v}^{\fl{k-1}},
\end{equation}
for all $0 \leq k \leq \Nt$ and  $\displaystyle{ M_{(\alpha,\cdot)}^{\fl{k}} := (\vert L_{(\alpha,\cdot)}^{\fl{k}}\vert +\vert R_{(\alpha,\cdot)}^{\fl{k}}\vert)^3}$. 

In particular,  $\displaystyle{\vert \U_{k}_v(X,\alpha^{\fl{k}}) \vert \leq(1+\dtu) M_{\alpha,v}^{\fl{k-1}}}$ holds  for all $0 \leq k \leq \Nt$.
\end{Proposition}
The previous results will be mostly used in the following way.
\begin{Corollary}\label{cor:GlobalODEsystem} 
Proposition \ref{prop:GlobalODEsystem} still holds with $L_{\alpha,v}^{\fl{k}}$ and $R_{\alpha,v}^{\fl{k}}$ substituted respectively by $L_{\alpha}^{\fl{k}}$ and $R_{\alpha}^{\fl{k}}$, where 
\begin{align*}
L_{\alpha}^{\fl{k}} := \inf_{-1\leq n \leq k}\left(\min_{1\leq v \leq \Nu}\left\{\alpha_v^{\fl{n}},0\right\}\right)\quad R_{\alpha}^{\fl{k}} := \sup_{-1\leq n \leq k}\left(\max_{1\leq v \leq \Nu}\left\{\alpha_v^{\fl{n}},1\right\}\right).
\end{align*}
In this case, \eqref{prop:GlobalODEsystemIneq} reads as $$(L_{\alpha}^{\fl{k-1}} - \dtu M_{\alpha}^{\fl{k-1}})\bm{1} \leq \U_{k}(X, \alpha^{\fl{k}}) \leq (R_{\alpha}^{\fl{k-1}} + \dtu M_{\alpha}^{\fl{k-1}})\bm{1},$$
with $M_{\alpha}^{\fl{k}} := (\vert L_{\alpha}^{\fl{k}}\vert +\vert R_{\alpha}^{\fl{k}}\vert)^3$.
\end{Corollary}
A  proof of this result is found in the Appendix \ref{app:proofs}.  It is important to observe  that these  bounds do not depend on $\Nt$, and may vary for different epochs: the result does not say how the invariant region evolves throughout the training process, and in fact it is not clear whether there exists such an  invariant region  for  trainable weights. Nevertheless,  training relies on  Gradient Descent, thereby  a synergy between the $\ell^{\infty}$-norm growth of $\U_{\cdot}(X,\alpha^{\fl{\cdot}})$ and that of  $W_{u}$  is expected. 
\subsection{Bringing in the phase equation, coupling both models, and further model compression}\label{sec:coupling_in}
We finally explain why \eqref{motivation} is made up of two systems of equations. It turns out that, unfortunately,  just using \eqref{fullmodela} for learning is doomed to failure, a fact that we explain with the help of the next result. (The proof is found in the appendix.) 

\begin{Lemma}[Monotonicity yields ``non-learnable'' classes]\label{lemma:monotonicity}
Let  $X, \widetilde{X} \in \R$ be two individuals satisfying the normalization condition \eqref{normalizationcondition}. Fix  $\displaystyle{0\leq \dtu < \frac{1}{10}}$, and consider \eqref{EqForu} with $\Nu = 1$. The discrimination rule   \eqref{TargetSpace:discriminantV2} takes the form $\displaystyle{h(X_{(i)}) = 1}$, if $\displaystyle{\U_{\Nt}(\widetilde{X};\alpha^{\fl{\Nt-1}})\geq \frac{1}{2}}$, and $h(X_{(i)}) = 0$ otherwise. Then, whenever $X \geq \widetilde{X}$, we have
$\displaystyle{\U_{n}\left(X;\alpha^{\fl{n}}\right) \geq \U_{n}\left(\widetilde{X};\alpha^{\fl{n}}\right), }$
for all $0 \leq n\leq \Nt.$
\end{Lemma}
The consequences of this result can be seen in an illustrative example. For a fixed  for $\gamma^* \in (0,1)$, let's label the data as  $ Y_{(i)} = \mathbbm{1}_{\{z\leq\gamma^*\}}(X_{(i)})$, and take $(X,Y)$ and $(\widetilde{X},\widetilde{Y})$ satisfying  $X > \gamma^* \geq \widetilde{X}$. Under these assumptions we claim that at least one of them is  wrongfully classified. Indeed, assume that both individuals  are correctly classified. Labeling and features' properties imply that  $Y = 0$ and $\widetilde{Y} = 1$, hence we must have  $\U_{\Nt}(X;\alpha^{\fl{\Nt-1}}) <\frac{1}{2} \leq \U_{\Nt}(\widetilde{X};\alpha^{\fl{\Nt-1}})$. However, since $X > \widetilde{X}$, this violates Lemma \ref{lemma:monotonicity}. 

In a few words, a model that consists of \eqref{EqForu} only has a flaw: it strongly depends on label assignment. 

Equation \eqref{EqForp} is introduced to repair this issue. The fixing is easily  motivated by considering  the 1D case in the continuum setting \eqref{ODEModel1:continuum}: first, we construct two linear maps,
\begin{equation*}
\mathcal{S}_1^{(0)}(u) = u, \quad \mathcal{S}_1^{(1)}(u) = 1- u, \quad u \in \R.
\end{equation*}
Observe that  $\mathcal{S}_1^{(1)}(0) = 1$ and $\mathcal{S}_1^{(1)}(1) = 0$; in other words, $\mathcal{S}_1^{(1)}(\,\cdot\,)$ flips the interval $[0,1]$, whereas $\mathcal{S}_1^{(0)}(\,\cdot\,)$ is simply the identity map. Such properties convey all that is needed: we choose  $\mathcal{S}_1^{(0)}(\,\cdot\,)$ when no relabeling is needed, otherwise we choose $\mathcal{S}_1^{(1)}(\,\cdot\,)$ and flip the labels, applying the previous model to $\mathcal{S}_1^{(1)}(\U_{\cdot};\alpha^{\fl{\cdot}})$. In practice,  figuring out when to use either of these maps involves the  construction of a homotopy,
\begin{equation}\label{01MapsConvexified}
\mathcal{S}_1^{(p)}(u) := (1- p)\,\mathcal{S}_1^{(0)}(u) + p\,\mathcal{S}_1^{(1)}(u),
\end{equation}
reducing matters to that of ``learning'' the homotopy parameter $p$.\footnote{Throughout the numerics the value of $p$ may scape the range $[0,1]$, but in terms of modeling it does what is intended, for optimization in $\beta$ is also in place. Similar maps are used in some gated RNN architectures; cf.  \citep[Chapter 10.10]{DeepLearning}.} The type of nonlinearity found in \eqref{ODEModel1:continuum} can be used for such purpose, namely, by setting  
$\displaystyle{\frac{d p}{dt} = f(p;\beta) := p(1-p)(p-\beta)}$, with  $\displaystyle{p(0) = \frac{1}{2}}$.
Now, $p= p(\cdot)$ has the desired properties as long as we learn the parameter $\beta$. In the end, going back to the continuum model, fixing the labels consists in obtaining $\displaystyle{\lim_{t\to \infty}\mathcal{S}_1^{(p(t))}\left(u(X_{(i)},t;\alpha)\right) = Y_{(i)}},$
which is achieved depending on the asymptotic behavior of $p(t)$ as $t \to \infty$. 

A multidimensional version of \eqref{01MapsConvexified}, more necessary to our goals,  is given by
\begin{equation}\label{01MapsConvexified_n_dim}
\R^{\Nu}\times \R^{\Nu} \ni (p,u)\mapsto \mathcal{S}_{\Nu}^{(p)}(u) := (\bm{1}-p)\oast\mathcal{S}_{\Nu}^{(\bm{0})}(u) + p\oast\mathcal{S}_{\Nu}^{(\bm{1})}(u).
\end{equation}
where both $\mathcal{S}_{\Nu}^{(\bm{0})}(u) = u$ and  $\mathcal{S}_{\Nu}^{(\bm{1})}(u) = \bm{1} - u$ are maps from $\R^{\Nu}$ to itself. 

After all this discussion, still,   more model compression can be attained.

\paragraph{Weight sharing within layers II.}  
Given elements $(X_{(i)}, Y_{(i)})$ in a data set, we would like to adjust parameters in Equation \eqref{fullmodel} in such a way that $\mathcal{S}_{\Nu}^{(p)}\left(\U_{\Nt}\left(X_{(i)};\alpha\right)\right)$ ``approximates'' $Y_{(i)}\bm{1}$. Since the role of  $p \in \R^{\Nu}$ in \eqref{01MapsConvexified_n_dim} is that of fixing label dependency, it will be taken by Equation \eqref{fullmodelb}. Hence, an immediate candidate consists of taking $p = \P_{\Nt}\bm{1}_{\Nu} \in \R^{\Nu}$, where $\P_{\cdot}$ evolves according to \eqref{fullmodelb}, with $\Np = 1$.

Among several other possible constructions of $p$, one that we adopt in this paper relies on  weight sharing. We once again use linear parameterizations: first, let  $ \U_{\cdot}$ be a solution to \eqref{EqForu} in such a way that the relation \eqref{BasisMatrix} holds. Recall from  Lemma \ref{BasisMatrix:decop} that $\U_{\cdot}_m$ depends only on columns  $b$  of $\mathscr{B}_u$ where   $m \in \mathrm{supp}(b)$.  Thanks to  this  ``evolutionary independence'', one can associate a unique 1D equation of the form \eqref{EqForp} to each of these columns, each with a similar role as that of $p$ in \eqref{01MapsConvexified}. Altogether, that would require coupling \eqref{EqForu} with a system of equations of the form \eqref{EqForp}, with  $\Np = \Npt$ variables.

We use \eqref{01MapsConvexified_n_dim} to  summarize these two cases.

\begin{Definition}[Subordinate and non-subordinate models]\label{def:SubNonSub} Given $1 \leq \Npt \leq \Nu$ and a basis matrix $\mathscr{B}_{\text{sub}}$ as in \eqref{BasisMatrix}. Then, two constructions are available:
\begin{enumerate}[label=(\roman*), ref=\theTheorem(\roman*)]
\item \textbf{(Non-subordinate phase)} Set $\Np :=1$ (hence, $\P_{\cdot}\in \R$) and $\mathscr{B}_{\text{sub}} = \bm{1}_{\Nu}  \in \R^{\Nu \times 1}$. Define
\begin{equation*}
\widetilde{\P_{\Nt}} := \mathscr{B}_{\text{sub}}\cdot \P_{\Nt} = \P_{\Nt}\bm{1}_{\Nu} \in \R^{\Nu},
\end{equation*}
where $\P_{\cdot}$ evolves by \eqref{fullmodelb}.
\item \textbf{(Subordinate phase)} Set $\Np :=\Npt$ (hence, $\P_{\cdot}\in \R^{\Npt}$). Define
\begin{equation*}%
\widetilde{\P_{\Nt}} := \mathscr{B}_{\text{sub}}\cdot \P_{\Nt} =\sum_{j = 1}^{\Npt} \P_{\Nt}_j e_{\pi_j^*}\in \R^{\Nu}, \quad \text{for} \quad \P_{\Nt}= \begin{bmatrix}
\P_{\Nt}_1\\
\vdots\\
\P_{\Nt}_{\Npt}
\end{bmatrix} \in \R^{\Npt},
\end{equation*}
where $\P_{\cdot}$ evolves by \eqref{fullmodelb}.
\end{enumerate}
\end{Definition}
Thus, for any chosen subordination, we couple \eqref{EqForu} with the system \eqref{EqForp} (with $\Np$ variables). Last,  we make use of \eqref{BasisMatrix} once more, parameterizing $\beta^{\fl{\cdot}} = \mathscr{B}_p W_p^{\fl{\cdot}}$; this time,  $ \mathscr{B}_p= \Id_{\Np}$.

Using this construction we are now ready to use the model \eqref{motivation} after combining it with a cost function,
\begin{equation}\label{Cost_3-1}
\widetilde{\mathrm{Cost}}_{\mathscr{D}} = \frac{1}{2\Nd}\sum_{i = 1}^{\Nd}\left\vert \mathrm{Mean}\left(\mathcal{S}_{\Nu}^{(\widetilde{P^{\fl{\Nt}}})}\left(\U_{\Nt}\left(X_{(i)};\alpha\right)\right)\right)- Y_{(i)}\right\vert_{\ell^2(\R^{\Nu})}^2,
\end{equation}
which is then optimized for $\alpha^{\fl{\cdot}}= \mathscr{B}_uW_u^{\fl{\cdot}}$ and $\beta^{\fl{\cdot}}= \mathscr{B}_pW_p^{\fl{\cdot}}$ (not shown, but implicit in the dynamics of \eqref{EqForp}). We highlight that  the cost function introduces interaction between  parameters $\alpha^{\fl{\cdot}}$ and $\beta^{\fl{\cdot}}$.  

\begin{figure}[htb]
\centering
\includegraphics[width=\textwidth, trim={0 0 0 0}]{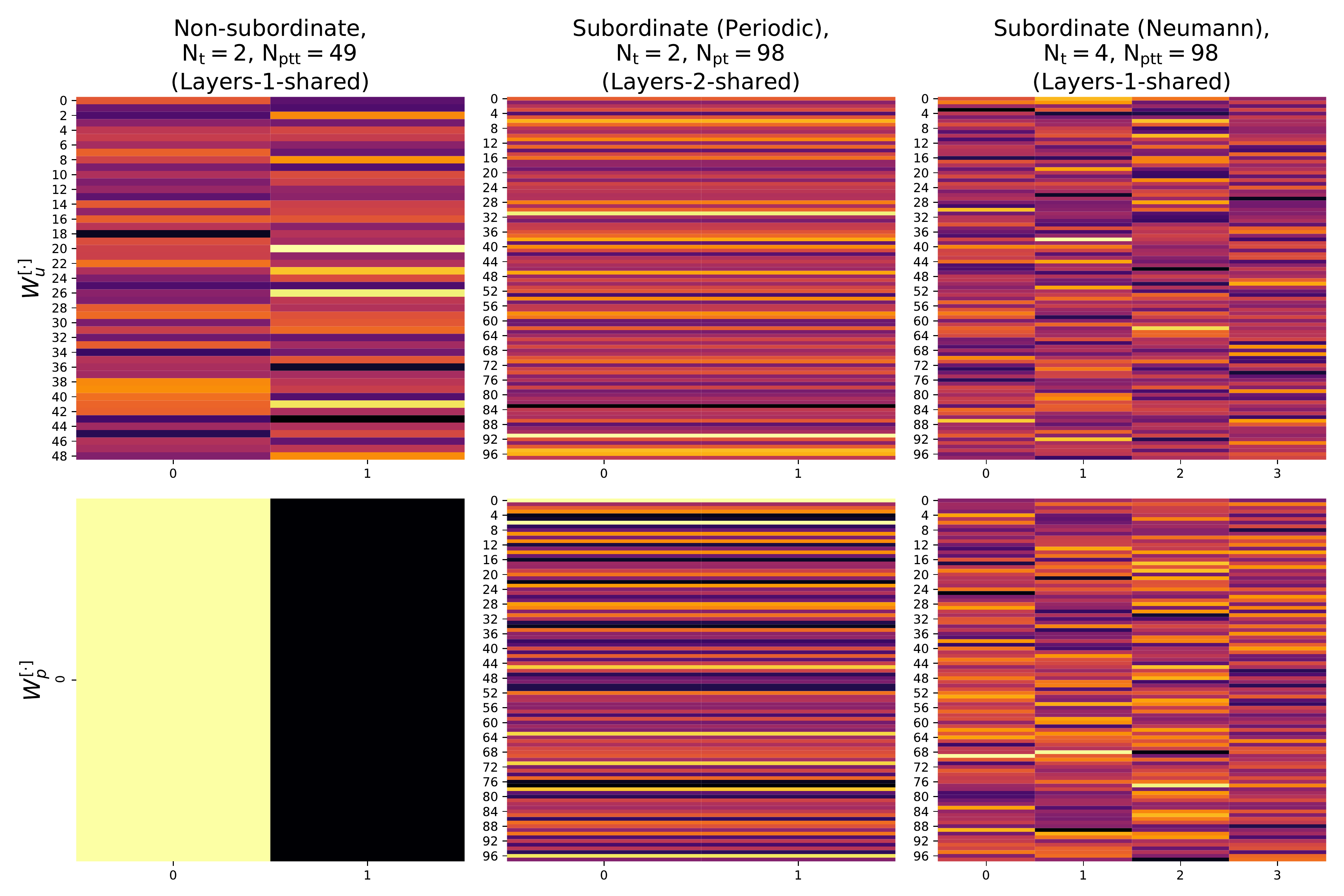}
\caption{Heatmaps of trainable weights for the PSBC with different architectures. \label{fig:InitializationTypesV2}}
\end{figure}

Different types of layers are illustrated in Figure \ref{fig:InitializationTypesV2}.

\begin{Remark}[Model parallelization]\label{rmk:parallelization}
In real applications of ML, due to the growing size of data sets, nearly all models rely on some sort of data parallelization (processing data in parts) using techniques like (minibatch) Stochastic Gradient Descent. On the other hand, achieving model parallelization (in which the model itself is split in fully independent parts) is much more complicated and an intense topic of research, with important practical implications to Engineering and Computer Science. 

It turns out that model parallelization can be obtained whenever $\ep =0$  using canonical basis matrices for  $\mathscr{B}_{\text{u}}$,  $\mathscr{B}_{\text{p}}$, and  $\mathscr{B}_{\text{sub}}$, as described above. In this way, due to Lemma \ref{BasisMatrix:decop},  features are forward propagated by $\U_{\cdot}$ and $\P_{\cdot}$ as $\Npt$ decoupled problems. Moreover, optimization splits into $\Npt$ smaller instances, that can be minimized independently. 
\end{Remark}
%
\subsection{Forward propagation $(\ep \geq 0)$ --- existence of dynamics}\label{sec:FullModel}
In light of Lemma \ref{BasisMatrix:decop}, the use of basis matrices constructed as in Remark \ref{rem:BasisMatrix} imply  that  features do not ``interact'' during forward propagation, which can be problematic because correlations are not fully captured until the cost function is evaluated.  We can mitigate this by  adding to \eqref{SystemGdFlow} a  penalization term of the form $\displaystyle{\frac{\ep}{2}\sum_{m=1}^{\Nu}\left\vert \U_{\cdot}_m - \U_{\cdot}_{m+1} \right\vert^2}$,  with due adjustments to account for boundary conditions;  the strength of this penalization is controlled by a diffusion parameter $\ep$.  Evidently, we do not add a similar term to  $\P_{\cdot}$'s equation, whose main goal is to fix  any possible label dependency. 

At first we limit ourselves to  homogeneous boundary conditions of Neumann type,
\begin{align*}
U_{0}^{\fl{n}} := U_{+2}^{\fl{n}}, \quad \text{and} \quad U_{\Nu+1}^{\fl{n}} := U_{\Nu-1}^{\fl{n}}, \quad \forall n\in \mathbb{N}.
\end{align*}
The connection between the PSBC and the feedforward network description in Figure \ref{fig:fwdprop} becomes more pronounced if we rewrite \eqref{fullmodela} in vectorial form, 
\begin{equation}\label{discretemodelvectorial}
\mathrm{L}_{\Nu}\U_{n+1} = \U_{n+1} - \ep^2 \, \Dd_{\Nu}\U_{n+1}= \U_{n} + \dtu\,f(\U_{n};\alpha^{\fl{n}} ),
\end{equation}
where 
\begin{equation}\label{DifferenceMatrixNeumann}
\begin{split}
\mathrm{L}_{\Nu} := \Id_{\Nu} -\ep^2 \, \Dd_{\Nu}, \quad \text{with} \quad 
\Dd_{\Nu} = \begin{bmatrix} 
-2 & 2 & 0 &\dots &0\\
1 & -2 & 1& \dots &0 \\
0 & 1 & -2& \dots &0 \\
\vdots & \ddots & \ddots & -2&1 \\
0 &  &0 & 2 & -2 
\end{bmatrix} \in \R^{\Nu \times \Nu}.
\end{split}
\end{equation}
A similar formula holds for \eqref{fullmodelb}, with $\mathrm{L}_{\Np} := \Id_{\Np}$. A direct application of Gershgorin's Theorem  implies that $\mathrm{L}_{\Nu}$ is invertible for all $\dtu\geq 0$, $\ep \geq 0$ \citep[Theorem 20.12]{linear_algebra-dym}. Therefore, \eqref{discretemodelvectorial} reads as
\begin{equation}\label{discretemodelexplicit}
\U_{n+1} = \mathrm{L}_{\Nu}^{-1}\cdot \left(\U_{n} + \dtu\,f(\U_{n};\alpha^{\fl{n}} )\right).
\end{equation}
Now, writing 
$Z^{\fl{n}} := \left(\begin{array}{c}
\U_n(X,\alpha^{\fl{n}})\\
\P_n\left(\frac{1}{2}\bm{1},\beta^{\fl{n}}\right)
\end{array} \right),$ 
with 
$\alpha^{\fl{n}} = \mathscr{B}_u\cdot W_{u}^{\fl{n}}$ and  $\beta^{\fl{n}} = \mathscr{B}_{p}\cdot W_{p}^{\fl{n}}$,
we define the activation function $\sigma^{\fl{n}}(\cdot, \cdot, \cdot, \cdot)$ as a map from $\R^{\Nu}\times \R^{\Nu}\times\R^{\Np}\times\R^{\Np}$ to itself,
\begin{equation}\label{GeneralFwdhierarchicalAsvec:diffusive}
\begin{split}
\sigma^{\fl{n}}(U, W_{u},P, W_{p}) := \left(\begin{array}{c}
        (\mathrm{L}_{\Nu})^{-1}\cdot\left( U + \dtu\,  f(U,W_{u})\right) \\
        P + \dtp \, f(P,W_{p})
        \end{array}
\right).
\end{split}
\end{equation}
Thus, we can write the discrete evolution \eqref{fullmodel} using the general framework presented in \eqref{fwdgeneral},
\begin{equation}\label{GeneralFwdhierarchicalAsvec}
\begin{split}
Z^{\fl{n+1}} = \left(\begin{array}{c}
\U_{n+1}\\
\P_{n+1}
\end{array}\right) = 
\sigma^{\fl{n}}\left(\U_n, W_{u}^{\fl{n}},\P_n, W_{p}^{\fl{n}}\right) = \sigma^{\fl{n}}\left(Z^{\fl{n}}, W^{\fl{n}}\right).
\end{split}
\end{equation}
It is also possible to study the PSBC model with periodic boundary conditions (when $\Nu \geq 3$), taking 
$U_{0}^{\fl{n}} := U_{\Nu}^{\fl{n}}$, and $U_{\Nu+1}^{\fl{n}} := U_{1}^{\fl{n}}$, for all $n\in \mathbb{N}$.
In such case a different diffusion matrix $\Dd_{\Nu}^{ (\mathrm{per})}$ is obtained, but  all the results related to the Invariant Region Enforcing Condition remain valid (mainly because  Maximum Principles still hold; see Appendix \ref{app:MP}).

Proposition \ref{prop:GlobalODEsystem} and Corollary \ref{cor:GlobalODEsystem} can be generalized to contemplate the dynamics in \eqref{fullmodel}.

\begin{Proposition}[Global existence of forward propagation --- general case]\label{prop:Globalpde}
Let $\Nt \in \N$ and $\U_{0} :=X\in [0,1]^{\Nu}$. Given trainable weights 
$\left(\alpha^{\fl{n}}\right)_{0\leq n \leq \Nt-1} \in \R$ and $\left(\beta^{\fl{n}}\right)_{0\leq n \leq \Nt-1}$, augmented by $\alpha^{\fl{-1}} := \bm{1}\in \R^{\Nu}$ and $\beta^{\fl{-1}} := \bm{1}\in \R^{\Np}$,  define for any $-1 \leq k \leq \Nt-1$ the quantities
\begin{align*}
\begin{split}
L_{\alpha}^{\fl{k}} := \inf_{-1\leq n \leq k}\left(\min_{1\leq m \leq \Nu}\left\{\alpha_m^{\fl{n}},0\right\}\right), \quad R_{\alpha}^{\fl{k}} := \sup_{-1\leq n \leq k}\left(\max_{1\leq m \leq \Nu}\left\{\alpha_m^{\fl{n}},1\right\}\right), 
\end{split}
\end{align*}
and, similarly, $L_{\beta}^{\fl{k}}$ and $R_{\beta}^{\fl{k}}$. Assume that 
$\displaystyle{-\infty < L_{\gamma}^{\fl{\Nt-1}}}$ and $\displaystyle{R_{\gamma}^{\fl{\Nt-1}}<+\infty}$, hold for  $\displaystyle{\gamma \in \{\alpha, \beta\}},$ 
and that the Invariant Region Enforcing Condition \eqref{IREC} applies as
\begin{align*}
0\leq \dtu\leq \frac{1}{\sqrt{3}\left(\vert L_{\alpha}^{\fl{\Nt-1}}\vert + \vert R_{\alpha}^{\fl{\Nt-1}}\vert \right)^2},  
\end{align*}
with similar conditions imposed on  $\dtp$.

Then, with either Neumann or Periodic boundary conditions,  the sequence $\U_k(X, \alpha^{\fl{k}})$ and $\P_k(\frac{1}{2}\cdot\bm{1},\beta^{\fl{k}})$ obtained using \eqref{GeneralFwdhierarchicalAsvec} remains bounded for all $0 \leq k \leq \Nt$.
More precisely, defining $M_{\alpha}^{\fl{k}} := \vert L_{\alpha}^{\fl{k}}\vert +\vert R_{\alpha}^{\fl{k}}\vert$, for any $0 \leq k \leq \Nt$ it holds that
\begin{equation}\label{prop:GlobalpdeIneq}
\left(L_{\alpha}^{\fl{k-1}} - \dtu M_{\alpha}^{\fl{k-1}}\right) \bm{1} \leq \U_{k}(X, \alpha^{\fl{k}}) \leq \left(R_{\alpha}^{\fl{k-1}} + \dtu M_{\alpha}^{\fl{k-1}}\right)\bm{1}.
\end{equation}
Similar bounds hold true for $\P_k(\frac{1}{2}\cdot\bm{1}, \beta^{\fl{k}})$, $L_{\beta}^{\fl{k-1}}$, $R_{\beta}^{\fl{k-1}},$ and $ M_{\beta}^{\fl{k-1}}$, $\dtp$ replacing, respectively, $\U_k(X, \beta^{\fl{k}})$, $L_{\alpha}^{\fl{k-1}}$, $R_{\alpha}^{\fl{k-1}},$  $ M_{\alpha}^{\fl{k-1}}$, and $\dtu$ in \eqref{prop:GlobalpdeIneq}.
\end{Proposition}
%
%
\section{The PSBC applied on the MNIST database}\label{sec:MNIST}
We finally apply the PSBC model to several classification problems using the MNIST database. There are several hyperparameters that can be chosen in many different ways, a myriad of  choices that we do not intend to exhaustively cover,  let alone qualitatively study. Instead, we wish to  understand how hyperparameters impact the model's accuracy, possibly degrading or improving it. Therefore no exhaustive search for their best combination was intended. 

Four different aspects of the PSBC and its model compression qualities are investigated: 

\begin{enumerate}[label=(\roman*), ref=\theTheorem(\roman*)]
\item Different numbers of layers are considered: $\Nt\in \{1, 2, 4\}$. 
\item Different layers-$k$-shared architectures are tested, for $k \in \{1, \Nt\}$. 
\item Different  parameterization cardinalities $\Npt$ are considered, with $\Npt \in \left\{\lfloor \frac{784}{k}\rfloor, \; \text{for}\; 1\leq k \leq 5\right\}$.
\item  Different Boundary conditions,  either Neumann or Periodic, are tested. Furthermore, the diffusion term is varied in $\ep \in\left\{0\right\} \cup \left\{\frac{1}{2^k}, \; \text{for} \;  0\leq k\leq 4\right\}$. Boundary conditions are the same at $\ep =0$ (because no diffusion is present).
\end{enumerate}
The methodology for model selection, model assessment, data preprocessing, and  statistics are explained next. Further details about the MNIST database are found in the Appendix \ref{app:InitializationOptmHyper}.
\paragraph{Model selection and model assessment.}\label{app:model_selection}
Initially,  the data set is split into train-development-test parts \citep[Chapter 7]{HTFElements}. Model selection is performed  using $5$-fold cross validations on the train-development set, where the hyperparameter with highest averaged accuracy is the one picked. Afterwards, with the chosen parameter in hands, we perform model assessment:  the model is trained (again), but now  on the whole train-development set. Accuracy is then measured on the test set which, up to this point, has been untouched by the model \citep[Chapter 11]{Shalev-Understanding_ML}. 

\paragraph{Data preprocessing and normalization.}  Preprocessing is necessary because MNIST images are 2D, but the PSBC requires a 1D initial condition. This is circumvented by flattening the images (i.e., writing them as column vectors), a procedural step also used in other models, like  ANNs; see  further discussion in  Section \ref{discussion:comparison}. 

Initially, preprocessing aims  to fulfill the normalization conditions \eqref{normalizationcondition}. But just this is insufficient, for  MNIST images have a high number of pixels 0's and 1's, values that are  stationary points under \eqref{fullmodela}. Thus, for each pair of distinct digits $(a,b)  \in \{0, \ldots ,9\}$  we use a normalization map of the form
\begin{equation}\label{normalization_map}
\mathscr{N}_{(a,b)}(X) = 0.5 \bm{1}_{\Nu} + 0.5 (X - \mu_{\text{training}}^{(a,b)}).
\end{equation}
where $\mu_{\text{training}}^{(a,b)}$ represents the average of the training set associated with those same digits.  The construction is designed in such a way that the statistical average over elements in the training set is  $0.5 \bm{1}_{\Nu}$;  the reason is because $0.5$ is the  average value of each  weight during initialization. Furthermore, it holds that $\mathscr{N}_{(a,b)}\left([0,1 ]^{\Nu}\right)\subset [0,1 ]^{\Nu}$. 

We remark that the normalization map depends on the pair of digits and the training sample only, remaining the same in different model assessments.

\paragraph{Statistics.} For any given set of hyperparameters, models are fitted 5 times. This is done in order to assess  statistical properties of the model, for weights are initialized in a randomized fashion. See initialization details in the Appendix \ref{app:initialization}.

\subsection{Results for the PSBC}\label{sec:results}

We consider the  PSBC with $\ep =0$, $\Nt =2$, and $\Npt = 196$, applied to 45 distinct pairs of digits $(a,b)$ in  $\{0, \ldots, 9\}$, trained and tested on the subsets corresponding to those same digits. For each one of these digits, $5$ PSBC models are fitted under the model selection and model assessment guidelines  explained above.  Statistics for each of these tests are shown in the Table \ref{table:binaries}. 

\begin{table}[htbp]
\centering
\begin{tabular}{p{.3cm}|ccccccccc} &           1 &           2 &           3 &           4 &            5 &           6 &            7 &           8 &            9 \\
\midrule
0 &  0.99$_{(0.0)}$ &  0.94$_{(0.0)}$ &   0.93$_{(0.0)}$ &  0.96$_{(0.0)}$ &  0.91$_{(0.01)}$ &  0.93$_{(0.0)}$ &  0.98$_{(0.0)}$ &  0.92$_{(0.0)}$ &  0.95$_{(0.0)}$ \\
1 &          &  0.95$_{(0.0)}$ &   0.96$_{(0.0)}$ &  0.96$_{(0.0)}$ &   0.94$_{(0.0)}$ &  0.97$_{(0.0)}$ &  0.95$_{(0.0)}$ &  0.94$_{(0.0)}$ &  0.97$_{(0.0)}$ \\
2 &          &          &  0.92$_{(0.01)}$ &  0.97$_{(0.0)}$ &   0.95$_{(0.0)}$ &  0.95$_{(0.0)}$ &  0.96$_{(0.0)}$ &  0.94$_{(0.0)}$ &  0.96$_{(0.0)}$ \\
3 &          &          &           &  0.99$_{(0.0)}$ &  0.83$_{(0.01)}$ &  0.98$_{(0.0)}$ &  0.97$_{(0.0)}$ &  0.88$_{(0.0)}$ &  0.97$_{(0.0)}$ \\
4 &          &          &           &          &   0.95$_{(0.0)}$ &  0.97$_{(0.0)}$ &  0.96$_{(0.0)}$ &  0.97$_{(0.0)}$ &  0.84$_{(0.0)}$ \\
5 &          &          &           &          &           &  0.95$_{(0.0)}$ &  0.96$_{(0.0)}$ &  0.92$_{(0.0)}$ &  0.95$_{(0.0)}$ \\
6 &          &          &           &          &           &          &  0.99$_{(0.0)}$ &  0.98$_{(0.0)}$ &  0.99$_{(0.0)}$ \\
7 &          &          &           &          &           &          &          &  0.96$_{(0.0)}$ &   0.9$_{(0.0)}$ \\
8 &          &          &           &          &           &          &          &          &  0.95$_{(0.0)}$\\\bottomrule\end{tabular}

\caption{Accuracies  on  test set (for the corresponding individuals associated with those labels).  Parentheses values indicated standard deviation.  Statistics consider $5$ independent model assessments. Values were rounded to 2 digits.
\label{table:binaries}}
\end{table}

Of the 45 cases,  $3$ have average in the range  $[80\%, 90\%)$, $11$ in the range $[90\%, 95\%)$, $31$  in the range $[95\%, 100\%)$. A more detailed analysis of these classifiers is done with the help of confusion matrices, as carried out in the next section (after aggregation). In both scenarios the qualitative summary is the same: while for some pairs of digits the PSBC shows high accuracy with remarkably small number of false positives and false negatives, in other cases the results are simply poor. Obviously, one of the simplest ways to improve any of these models is by doing a more extensive grid search for hyperparameters (ours was in fact small, made up of about 100 different points). Instead, in  Section \ref{sec:pca} we go in a different direction, discussing how other ML techniques can be used to enhance the PSBC. 

\subsection{Constructing a multiclass classifier.}\label{sec:multiclass}

There are several possible approaches to generalize a binary classifier to a multiclass classifier. For illustration, we resort to combining classifiers by aggregation, followed by a  \textit{one versus one} technique \citep[Chapter 18.3.3]{HTFElements}. The binary classifiers used are those from Section \ref{sec:results}, hence we continue with hyperparameters  $\ep = 0$,  $\Nt = 2$, and $\Npt = 196$.  
\begin{figure}[htbp]
\centering
\includegraphics[width=\textwidth]{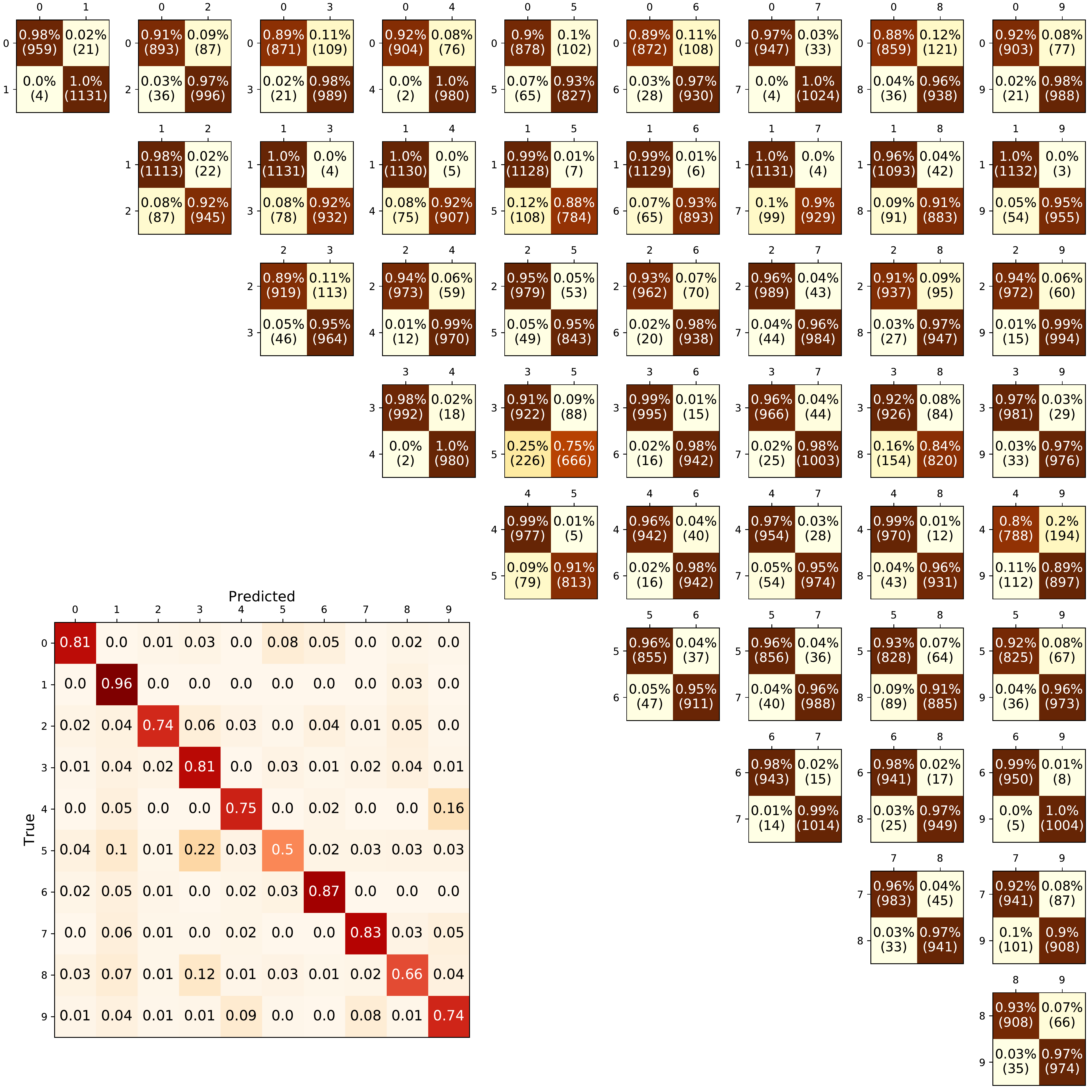}
\caption{
On the lower left, the confusion matrix of a multiclass classifier constructed with committees of PSBC (binary) classifiers (top, right). Each binary classifier's parameters were $\Nt = 2$, $\Npt = 196$, and $\ep = 0$. Model selection  is carried out by grid search for  learning rates, $\dtp$, $\dtu$. The best chosen models is trained 5 times on the training-development set. Afterwards, a final classifier is constructed using Ensemble learning with hard-voting. Accuracy values on the test set are those displayed. Accuracy of the multiclass classifier is 77.25\%. Values were rounded to 2 digits. \label{fig:multiclass}}
\end{figure}
 
We finally turn to the construction of the multiclass classifier. Initially, for each distinct pair of  digits  $(a,b)  \in \{0, \ldots ,9\}$ we combine the $5$ classifiers described in the previous section into  a single classifier using Ensemble Learning principles with hard voting \citep[Chapters 8.7]{HTFElements}. The resulting classifier - which we denote by $\mathbb{P}_{(a,b)}(\cdot)$ - also has an associated map $\mathscr{N}_{(a,b)}(\cdot)$ constructed as in \eqref{normalization_map}; observe  that  $\mathbb{P}_{(a,b)}(\cdot)$ varies over model assessments due to statistical variability of the classifiers, while $\mathscr{N}_{(a,b)}(\cdot)$ does not. Afterwards, multiclass prediction proceeds by majority voting.\footnote{\label{hard_voting_footnote} In details, it goes as follows. First, individuals  in the test set are min-max normalized to fit into the box $[0,1]^{\Nu}$. They are normalized again by $\mathscr{N}_{(a,b)}(\cdot)$ before classification using the model $\mathbb{P}_{(a,b)}(\cdot)$. Without loss of generality, we assume that, whenever $a<b$,  $a$ receives label $0$, while $b$ receives label $1$. Then, all  votes are computed as $\displaystyle{\mathbb{M}_{(a,b)}(\cdot) = 1- \mathbb{P}_{(a,b)}(\cdot)}$ if $a<b$, $\displaystyle{\mathbb{M}_{(a,b)}(\cdot) = \mathbb{P}_{(a,b)}(\cdot)}$ if $b<a$, and zero otherwise. Afterwards, votes are counted as $\displaystyle{\mathbb{C}_{a}(\cdot) = \sum_{b \neq a}\mathbb{M}_{(a,b)}(\cdot)}$ The predicted label is assigned by sampling uniformly among individuals with highest score, so as to avoid ties; that is,  sampling uniformly from   $\displaystyle{\{a|\mathbb{C}_{a}(X) = \max_{0\leq \gamma\leq 9}\mathbb{C}_{\gamma}(X)\}}$.}  

Results are shown in Figure \ref{fig:multiclass}. As one can see,  the accuracy obtained on the test set is not high: 77.25\% only. Evidently, there are many ways to improve this figure. For instance, one could begin by improving each one of these binary classifiers, whose accuracies vary considerably across different pairs of digits; some ideas are discussed and tested in Section \ref{sec:pca}. Another approach, one that takes the variability in classifiers' accuracies into account, pushes hard voting aside and assigns weights to voters instead; this can be achieved using Boosting methods \citep[Chapters 10 and 16]{HTFElements}. 

\subsection{Modeling the data manifold by changing the basis matrix.}\label{sec:pca} 
As pointed out in Remark \ref{BasisMatrix}, an obvious way to improve the model is by allowing the basis matrices $\mathscr{B}^{ \fl{\cdot}}$ to vary over layers. Evidently, in light of Lemma \ref{BasisMatrix:decop} such variations change the architecture of the network, distorting the approximation space where the prediction map is built. 

Faced with many possible choices for basis matrices, some intuition about the role of trainable weights can be useful. During training, for example: in light of Lemma \ref{lemma:monotonicity} ($\Nu = 1$), one can conclude that Equation \eqref{fullmodela} is hyperplane separating the set of initial conditions by proper adjustment in the values of $\alpha^{\fl{\cdot}}$ (and, whenever necessary, fixing labels by changing $\beta^{\fl{\cdot}}$). Even in a high-dimensional feature space we can still say that this is what is happening when canonical basis matrices are used  (thanks for Lemma \ref{BasisMatrix:decop}). 

\begin{figure}[htb]
\begin{subfigure}[b]{.5\textwidth}
\centering  
\includegraphics[width=\textwidth, trim = {0cm 1cm 0cm 1cm}]{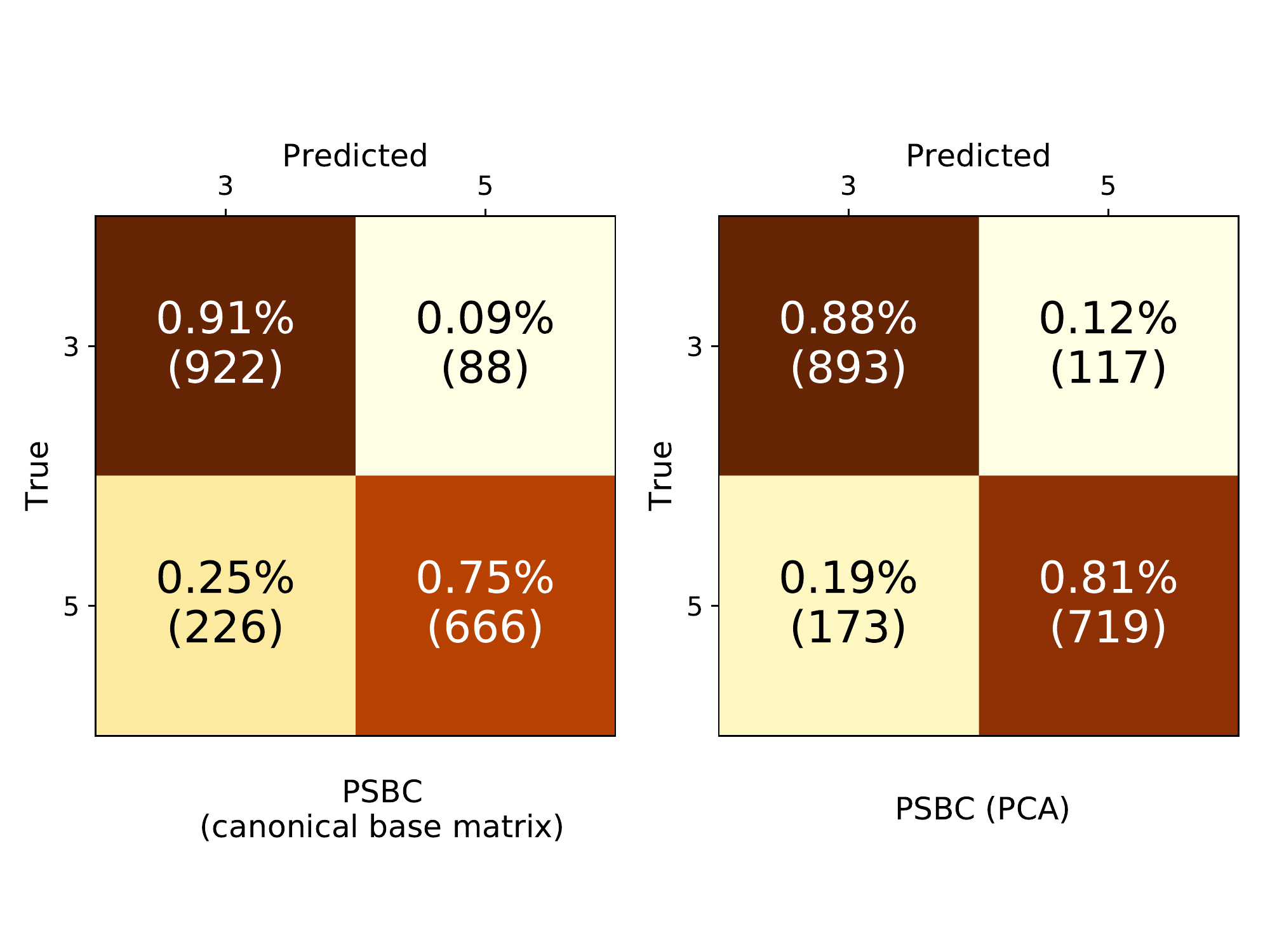}
\caption{Digits 3 and 5.\label{fig:a}}  
\end{subfigure}
\hfill
\begin{subfigure}[b]{.5\textwidth}
\centering
\includegraphics[width=\textwidth, trim = {0cm 1cm 0cm 1cm}]{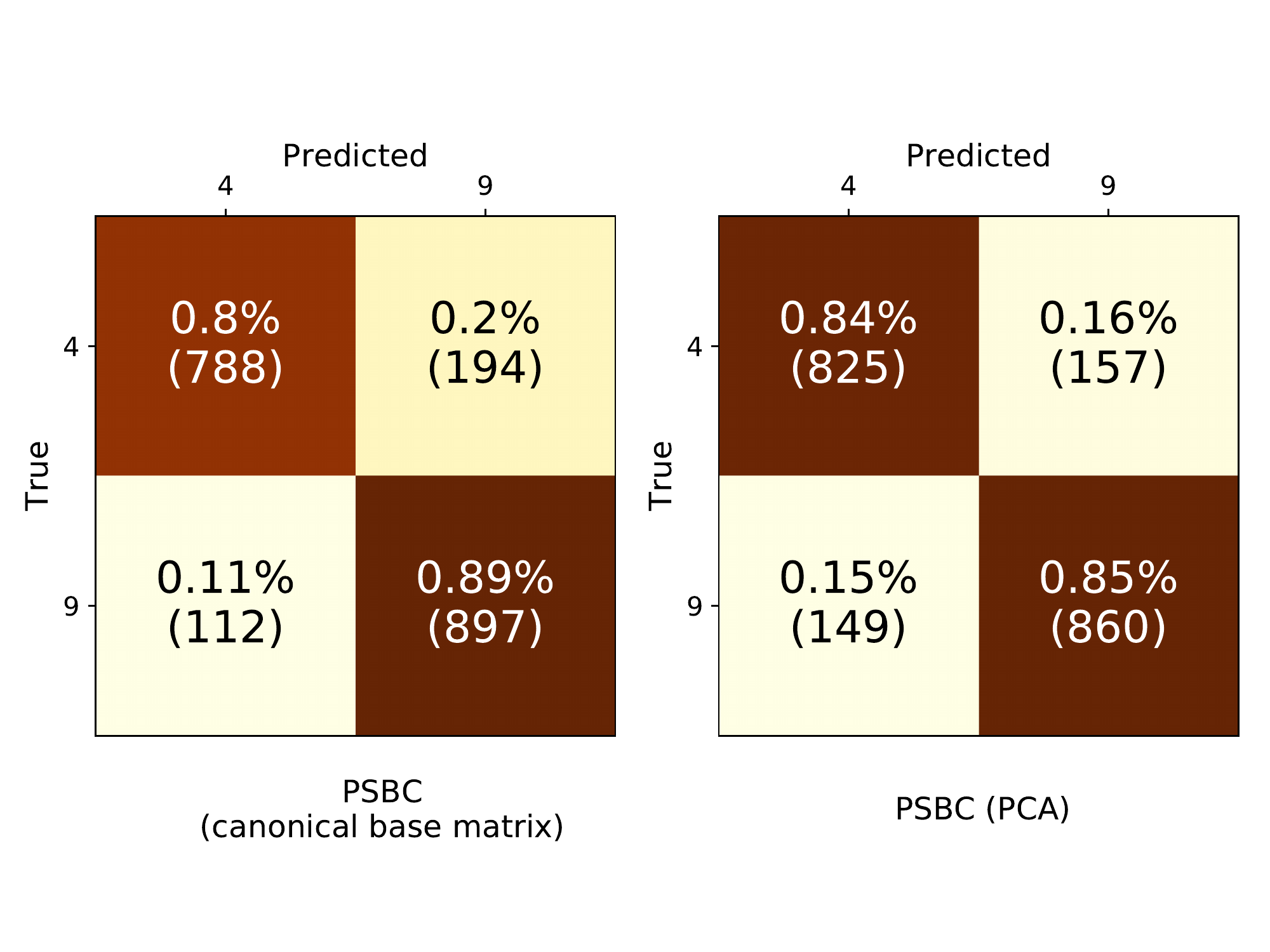}
\caption{Digits 4 and 9. \label{fig:b}}  
\end{subfigure}

%
%
%
%
%
%
%
 
\caption{
Confusion matrices comparing different models that were constructed using Ensemble Learning methods, with different base classifiers - PSBC models with  canonical base matrix or PSBC models with base matrices built using PCA. 
The confusion matrix for digits 3 and 5 is shown in  \ref{fig:a}, showing some improvement in accuracy, which 
goes from 83.49\% to 84.75\%, while F1 score jumps from 80.92\% to 83.22\%
On the other hand, the confusion matrix for digits 4 and 9 is shown in  \ref{fig:b}, but in this case the result is mixed, and seems to deteriorate: accuracy remains the same (84.63\%), while F1 scores shows a slight decrease, from  85.43\% to 84.9\%. Values were rounded to 2 digits.
\label{fig:PCA_examples}}
 \end{figure}

A less direct but maybe more  interesting approach consists of modeling the basis matrix on the data manifold. We did some studies in this regard, combining the PSBC with Principal Components Analysis (PCA) \citep[Section 14.5]{HTFElements}. In this case, $\mathscr{B}_u$ consists of the first $\Npt$ principal components associated with the trained data, properly centered for such computations.   As before,  we take $\mathscr{B}_p = \Id_{\Np}$, for a subordinate model ($\Np = \Npt$). Although the results improve those shown in Figure \ref{fig:multiclass}, the choice for $\mathscr{B}_{\text{sub}}$ holds with less motivation, for it is still constructed in the canonical way described in Remark \ref{rem:BasisMatrix}. The results displayed in Figure \ref{fig:PCA_examples} are interesting, indicating that the lowest scores among False Positive and False Negatives increase, while the highest values among False Negatives and False Positives decay, making predictions using  the PSBC with PCA more ``balanced'' than with the usual PSBC. In any case, these results are mostly illustrative, and more statistical evidence or theoretical understanding is required. Further technical details, mostly concerning the application of Invariant Region Enforcing Condition \eqref{IREC}, are discussed in Section \ref{sec:invariant}.

It must be said that when PCA is applied  neither sparsity nor disjointness of basis elements' support (principal components) are expected, properties which imply  model parallelization in the non-diffusive PSBC ($\ep =0$), according to Lemma \ref{BasisMatrix:decop}. In reality, and in spite of the results shown in this section, it is unclear what qualities sparse basis matrices offer to the PSBC. Nonetheless, it could be interesting to investigate the construction of basis matrices using techniques like Sparse Principal Components, with the caveat that Lasso methods foster sparsity, but not disjointness of basis' supports; cf \citep[Section 14.5.5]{HTFElements}. If model parallelization is indeed necessary, this search can be converted into a combinatorial optimization problem.

\subsection{Varying parameterization cardinality and viscosity rates --- a study for the digits ``0'' and ``1''.}\label{sec:viscosity_and_ptt_cardinality}
As we explain below, model selection of hyperparameters was carried out considering that  the model is computationally much cheaper at $\ep =0$, when no diffusion matrix multiplication is involved. For that reason we are specially interested in studying the impact of  viscosity on model's accuracy. 

Evidently, different parametrization cardinalities yield different models, where a different number of weights are optimized. Nevertheless, with all hyperparameters fixed but $\ep$, it is noteworthy that a non-diffusive $(\ep =0)$ and a diffusive $(\ep >0)$ models have the same number of trainable weights. Moreover, it is much cheaper and faster  to train the first than the latter. Combining these facts, it seems interesting to investigate whether hyperparameters chosen through model selection  in a non-diffusive setting ($\ep =0$) are still ``good'' hyperparameters for diffusive models ($\ep >0$), all other hyperparameters  fixed. Does this model selection strategy yield improvement - or deterioration  - of  models' accuracies?

With this question in mind, we apply model selection as follows: initially, hyperparameters $\Nt$, $\Npt$, and layer-$k$-shared  (for $k \in \{1, \Nt\}$) are fixed. Grid search is carried out at $\ep =0$ for  learning rates,\footnote{Each set of weights in \eqref{fullmodela} and \eqref{fullmodelb} allow for different learning rates.} $\dtu$, and $\dtp$; they amount to  4 hyperparameters  that are tuned through model selection, as explained earlier. The best hyperparameters are used for training the PSBC at all values of  $\ep \geq 0$ considered. In all the cases, model assessment (evaluation) uses the same training and test sets. 

\begin{figure}[htb]
\centering
\includegraphics[width=\textwidth, trim = {1cm 2.3cm 1cm 1cm}]{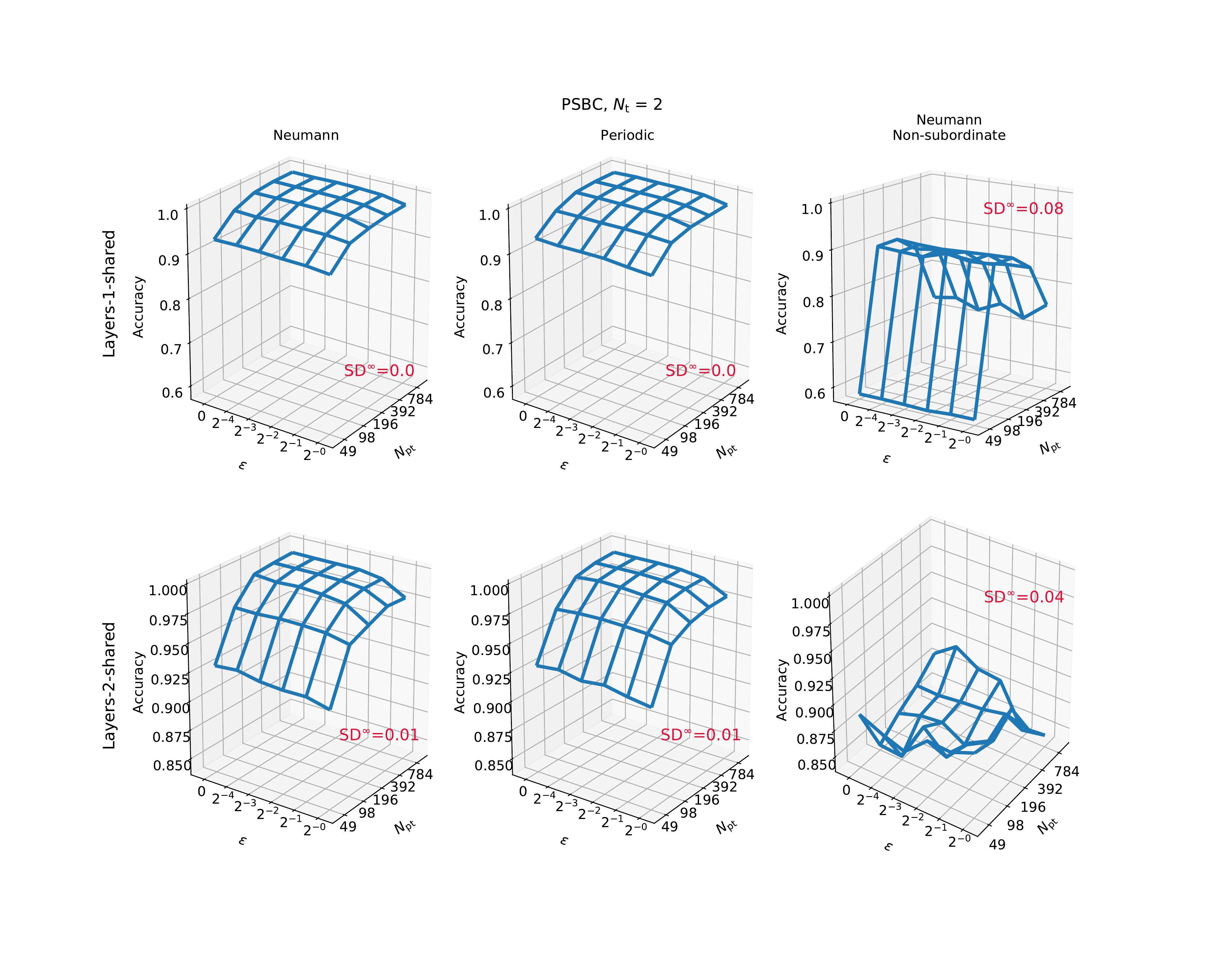}
\caption{Accuracy for different values of viscosities $\ep$ and parameterization  cardinalities $\Npt$.
The classifier is applied to the ``0'' and ``1'' subset of the MNIST database. Each point on these surfaces is computed as an average of 5 experiments, with an associated standard deviation; $\mathrm{SD}^{\infty}$ indicates the largest among them. For convenience the positive part of the  $\ep$ axis is in log-scale, whereas the $\Npt$ axis is off scale, represented as an equally spaced grid. Values were rounded to 2 digits. \label{PSBC_accuracy_Nt_2}}
\end{figure} 

Figure  \ref{PSBC_accuracy_Nt_2}  provides a comparison  between the model's behavior for different values of parameterization cardinality, as the diffusion term $\ep$ varies. Several observations are worthwhile mentioning. 

First, the type of boundary condition does not seem to affect the model's accuracy much. This is somehow expected due to flattening,   but should not discourage further investigation, especially those generalizing the model to 2D features; see discussion in Sections \ref{discussion:Periodicdiffusion_matrix} and \ref{discussion:multi_d}. Second, that diffusion does not seem to improve the model trained at $\ep =0$, being in fact detrimental to accuracy in both cases - Neumann and Periodic boundary condition; on the other hand, diffusion seems to improve accuracy in some models with Neumann, non-subordinate architecture. Third and last, that in the case of Periodic and Neumann boundary conditions with high parameterization cardinality ($\Npt \approx \Nu$), the cheapest PSBC with layers-$\Nt$-shared architecture (all layers are shared) perform as well as a PSBC with layers-1-shared (more expensive for optimization). Surprisingly, in some cases the classifier achieves on average more than 90\% accuracy even  when $\Npt =49$ the model, that is,   with less than 100 variables (layers-2-shared), a remarkable manifestation of PSBC's model compression qualities and an impossible milestone for ANNs with no weight sharing, where the minimum number of nodes is always bigger than the dimension of the input space. 

Similar figures, for $\Nt \in \{1,4\}$, are available in the Supplementary Material.
\subsection{The Invariant Region Enforcing Condition in practice.}\label{sec:invariant} 

\begin{figure}[htb]
\centering
\includegraphics[width=\textwidth, trim = {2cm 2cm 2cm 2cm}]{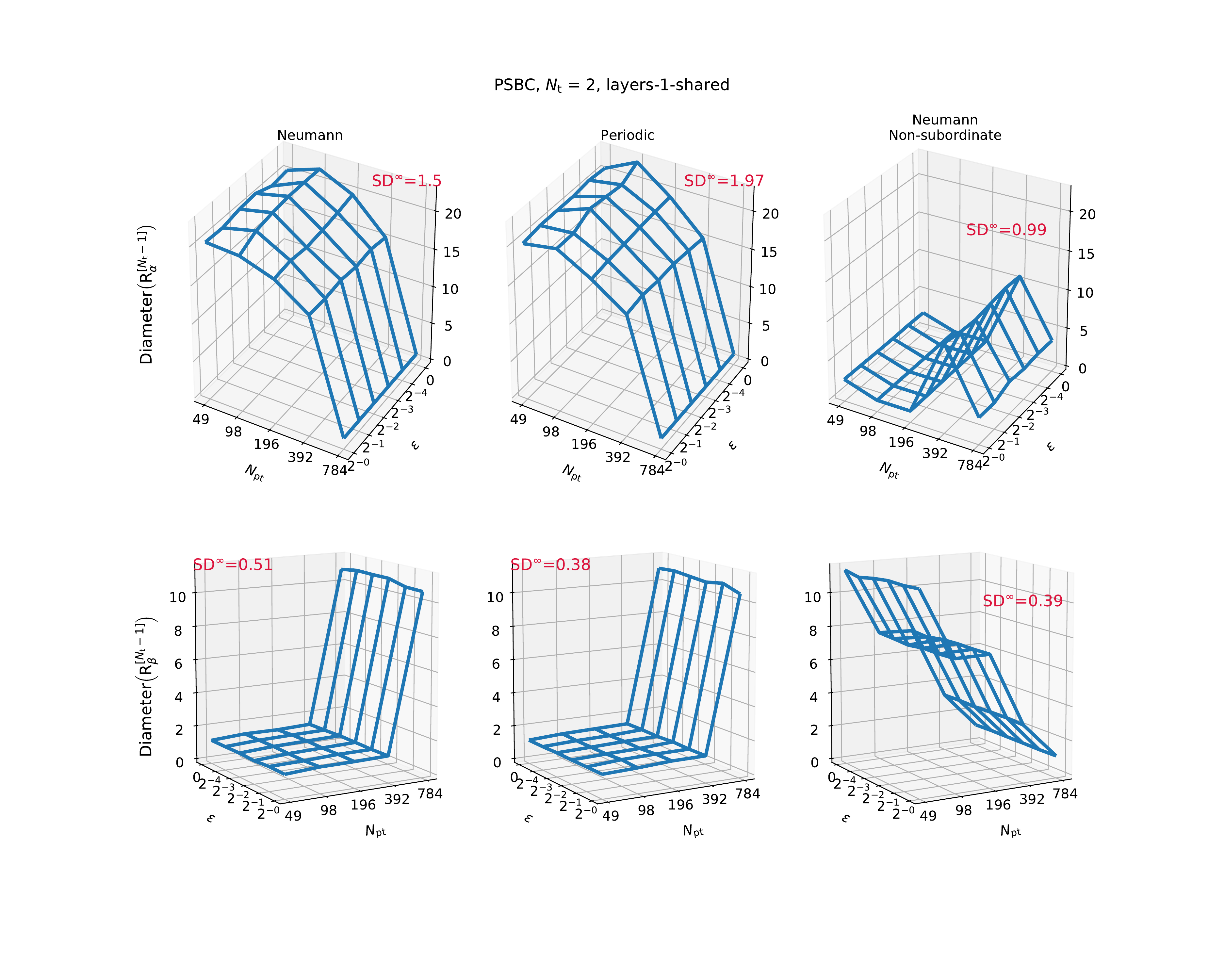}
\caption{Average values  attained by the diameter of the set $\mathcal{R}_{\alpha}^{\fl{\Nt -1}} :=\mathrm{conv}\left(\{0,1\} \cup_{m=0}^{\Nt-1}\{ \alpha^{\fl{m}}\}\right)$ at the best epoch (according to an Early Stopping criterion); $\mathcal{R}_{\beta}^{\fl{\Nt -1}}$ is defined similarly. 
Different values of viscosities $\ep$ and parameterization  cardinalities $\Npt$ have been tested. The classifier is applied to the ``0'' and ``1'' subset of the MNIST database. Each point on these surfaces is computed as an average of 5 experiments, with an associated standard deviation; $\mathrm{SD}^{\infty}$ indicates the largest among them. For the time evolution of these diameters through epochs, see the figures in the Supplementary Material and  also the the accompanying movie file  \citep[\red{Example\_layers\_snapshots.mp4}, about 0.5 Mb]{Bin_phase_github}. For convenience the positive part of the  $\ep$ axis is in log-scale, whereas the $\Npt$ axis is off scale, represented as an equally spaced grid.  Values were rounded to 2 digits.\label{fig:max_weights}}

\end{figure}

It is important to study the behavior of trainable weights of optimized models. In Figure \ref{fig:max_weights} we  analyze how different boundary conditions, diffusion, and parameterization cardinality affects the sup norm of the best models we obtained (whose accuracies are displayed in Figure \ref{PSBC_accuracy_Nt_2} in some cases; see also  the Supplementary Material).  

We must emphasize that the interval $\mathcal{R}_{\alpha}^{\fl{\Nt-1}} :=\mathrm{conv}\left(\{0,1\} \cup_{m=0}^{\Nt-1}\{ \alpha^{\fl{m}}\}\right)$ is built using elements on the range of the map \eqref{BasisMatrix}.  For this reason, when arbitrary basis matrices $\mathscr{B}_u$ are used, $\mathcal{R}_{\alpha}^{\fl{\Nt-1}} $  may not be easily recovered in terms of $W_u^{\fl{\cdot}}$'s entries, making computations using the PSBC slightly more expensive. Luckily, that's not an issue when canonical basis matrices are used; in such case, according to Remark \ref{rem:BasisMatrix}, we can rewrite 
$$\mathcal{R}_{\alpha}^{\fl{\Nt-1}} = \mathrm{conv}\left(\{0,1\} \cup_{m=0}^{\Nt-1}\{ W_u^{\fl{m}}\}\right).$$  
This property can be used to understand the $\ell^{\infty}$-norm of weights, since
$$\displaystyle{\max_{0\leq n \leq \Nt-1}\left\{ \Vert\alpha^{\fl{n}}\Vert\li , \, 1\right\}  \leq \mathrm{diam}(\mathcal{R}_{\alpha}^{\fl{\Nt -1}})  \leq 2 \max_{0\leq n \leq \Nt-1}\left\{ \Vert\alpha^{\fl{n}}\Vert\li , \, 1\right\}};$$   
bounds in terms of $\Vert W_u^{\fl{\cdot}} \Vert\li$ are immediate, thanks to Lemma \ref{BasisMatrix:param}. Similar statements are valid  in the case of $\mathcal{R}_{\beta}^{\fl{\cdot}}$, $\beta^{\fl{\cdot}}$, and $W_p^{\fl{\cdot}}$.

In Figure \ref{fig:max_weights} we study the behavior of trainable weights attained at best accuracy (according to Early Stopping).  It indicates that the diameters of $\mathcal{R}_{\alpha}^{\fl{\cdot}}$ and $\mathcal{R}_{\beta}^{\fl{\cdot}}$  grow at different rates, implying that $\dtu$ and $\dtp$ shrink towards zero with different rates in virtue of \eqref{IREC}. In fact, this observation is in good agreement with our numerical experiments; see the Supplementary Material.

Earlier simulations on the MNIST database have shown that  imposing the  constraint  $\dtu = \dtp$  in each epoch does not give good results. Furthermore, allowing $\dtu$ and $\dtp$ to assume their maximum value gives poor accuracy results, probably because in this way the model's hyperparameters vary too many times over different epochs. To circumvent such an issue, after each batch evaluation we use update $\dtu$ and $\dtp$ using the Invariant Region Enforcing Condition \eqref{IREC} in the form
$$ \dtu := \min\left\{\dt^*, \frac{1}{\sqrt{3} \, \mathrm{diameter}\left(\mathcal{R}_{\alpha}^{\fl{\Nt -1}}\right)^2} \right\},$$
with a similar definition holding for $\dtp$. Here, $\dt^*$  indicates $\dtu$'s initial value.  Further information about parameters initialization can be found in the Appendix \ref{app:InitializationOptmHyper}.
%
%
\section{Discussion, some extensions,  and open questions}\label{sec:further_results}
As we recall from Section \ref{sec:introduction},   binary classification  concerns the construction of  approximations $\widetilde{h}(\,\cdot\,)$ to an unknown hypothesis map $h(\cdot)$. Some of these methods can be motivated by biological mechanisms: for instance, the design and structure of CNNs bear similarity to how cortical neurons are scattered throughout the brain, while CNNs' functioning resemble cortical neurons orientation selectivity to visual stimuli. Likewise, the design of ANNs is reminiscent of the brain's networks \citep[Chapters 1 and 9.10]{DeepLearning}. In regard to the  PSBC, whenever $\Nu=1$ and $\ep =0$ its component  \eqref{fullmodela} can be interpreted as a discrete ODE modeling a nothing-or-all system, that is, a phenomenon of  binary nature, which resembles  neurons' reaction to stimuli: depending on a threshold (in this case, denoted by $\alpha(\,\cdot\,)$), stimuli are classified as inhibitory or excitatory. In terms of mathematical or physical modeling these resemblances are mostly metaphoric, even though,  to the author's knowledge, likening binary classification to phase separation processes seems to be new. 

Nonetheless, we have been inspired by many of these ideas (specially by those we oppose). First, by introducing a model that is not based on  brain-like phenomena we just reiterate our belief that such quality is simply unnecessary. Second, by likening binary classification to the process of phase separation in binary fluids. Third, by likening stable (asymptotic) patterns in Reaction-Diffusion Equations to the process of labeling in supervised learning. Last, by showing that forward propagation can be done without squashing nonlinearities, turning it into an initial boundary value problem that is amenable to mathematical techniques from the field of PDEs.
  
In the sequel we make a few remarks concerning the  PSBC.  The topics addressed  point out to directions in which the model can be improved, investigated, or extended in manners that we did not, at least in this first paper, exploited in depth. 

\subsection{On diffusion matrices and  boundary conditions}\label{discussion:Periodicdiffusion_matrix}

Besides different boundary conditions, other modifications in the diffusion matrix can be envisioned if one observes that the matrix $\Dd_{\Nu}$ represents  a first-order accurate finite difference scheme  for  the operator $\partial_x^2$ \citep[Chapter 3]{Strikwerda}. One could also consider  finite difference schemes of higher order of accuracy, to which new diffusion matrices are associated. 

We highlight that the matrix  $\mathrm{L}_{\Nu}$ in \eqref{discretemodelvectorial} is invertible for all $\ep^2\geq 0$ in both boundary condition cases, Neumann  or Periodic. Hence,   varying $\ep^2$  is legitimate and does not affect the Invariant Region Enforcing Condition, which  does not depend on $\ep$. It is therefore plausible, at least in principle,  that the parameter $\ep^2$ could also  be trained using backpropagation, in such a way that diffusion could be learned ``on-the-fly''. Nonetheless, the diffusion matrices considered here - $\Dd_{\Nu}$ and  $\Dd_{\Nu}^{\text{(per)}}$ - are too ``rigid'',  in the sense that they couple features with the same weights, regardless of their statistical correlation. It would be interesting to optimize the viscosity matrix as a trainable  weight while keeping its negative semidefinite structure. This approach would generalize the periodic case studied  here, and  can be achieved by considering a Cholesky factorization of the diffusion as $D = -V V^T$ and optimizing for the entries in $V$. 
 
\subsection{Extending the PSBC using a  multi-d diffusion operator}\label{discussion:multi_d}
A deeper analysis of  \eqref{DifferenceMatrixNeumann} indicates that the diffusion term $\ep$ controls the strength of interaction between features. This can be seen  by a Neumann series expansion,
\begin{equation*}
\mathrm{L}_{\mathrm{N}}^{-1} = (\Id_N - \ep^2 \Dd_{\mathrm{N}})^{-1} = \sum_{n = 0}^{\infty} \ep^{2n} \Dd_{\mathrm{N}}^n, 
\end{equation*}
which is valid for sufficiently small values of $\ep$ \citep[Chapter 7.7, Lemma 7.15]{linear_algebra-dym}.
One immediately concludes that $\mathrm{L}_{\mathrm{N}}^{-1}$ is diagonal dominant. Moreover, due to the band-limited structure of $\Dd_{\mathrm{N}}$ and its powers, it implies  that elements  of $\mathrm{L}_{\mathrm{N}}^{-1}$ decay polynomially in $\ep$ the farther away from the diagonal they are. 

The comments above lead us to a concerning conclusion: that the PSBC favors features interaction based on the proximity of their indexes. Flattening seems then unsatisfactory, in spite of high accuracy results obtained here. It should be noted that flattening does not affect models like  ANNs due to their fully connected layers.

Removing the PSBC's dependence on flattening is an important - and tangible - direction of improvement, which can be attained if one extends the model by changing \eqref{fullmodela} to a  multi-d spatial diffusion. Mathematically, this amounts to substituting the 1D Laplacian $\partial_x^2$ in \eqref{AC} by a 2D Laplacian  $\partial_x^2 + \partial_y^2$. As mentioned before, any discretization of the 2D Laplacian obeying the Maximum Principles as developed in  Appendix \ref{app:MP} and \ref{app:proofs} yields a model with  similar mathematical properties - Invariant Regions, Invariant Region Enforcing Condition etc. 

Overall, there is another reason for allowing more general diffusion matrices: learning boundary conditions rather than imposing them. Indeed, certain types of CNNs require padding of the input data with zeros, creating bias in the statistics close to the pictures' edge \citep[Chapter 9.5]{DeepLearning}. On the other hand, intuitively, it makes more sense to have boundary conditions that adapt statistical weights to features close to an edge. 

It is plausible that a combination of a  multidimensional diffusion model with the learning of general diffusion matrices (as discussed in Section \ref{discussion:Periodicdiffusion_matrix}) may not only result in better binary classifiers, but also in models that have considerable overlap with those developed and used in Computer Vision. 
%
\subsection{Invariant regions, growth of trainable weights in $\ell^{\infty}$-norm, and the enforced invariant region condition}\label{discussion:invariant_region}
  
We have shown in Propositions \ref{prop:GlobalODEsystem}, and \ref{prop:Globalpde} that  $\U_{\cdot}$ and $\P_{\cdot}$ do not blow-up during forward propagation under appropriate assumptions. These results are important to the extent that  they establish and secures one of the basic mechanisms used in the construction of the prediction map \eqref{classassignment}. %
Yet, there are many open questions about the asymptotic behavior of trainable weights. For instance,  it is unclear whether  an invariant region for trainable weights exists (in terms of hyperparameters and quantitative estimates on the data set $\mathscr{D}$ used for its training). Overall,  it is not known if either random initialization of trainable weights or the Invariant Region Enforcing Condition \eqref{IREC} are sufficient to ensure the existence of such an invariant region.

\subsection{Comparison with other Machine Learning models}\label{discussion:comparison}

We have pointed out many similarities between the PSBC model and classical ones, like  ANNs, CNNs, and RNNs. There are also some similarities when we compare the PSBC and CNNs. For instance, in the periodic case one case see the matrix $\mathrm{L}_{\mathrm{N}}^{-1}$ as a Toeplitz matrix that plays the role of a convolution in \eqref{discretemodelexplicit}  \citep[Chapter 9.1]{DeepLearning}. The comparison becomes more explicit when we consider the fully explicit model (as in the Footnote \ref{theta_model}),
\begin{equation}\label{discretemodelfully_explicit-theta}
 \U_{n+1} = \left(\Id_{\mathrm{\Nu}} +\ep^2 \Dd_{\Nu}^{ (\mathrm{per})}\right) \U_{n} + \dtu\,f(\U_{n};\alpha^{\fl{n}} ),
\end{equation}
where the matrix $\Id_{\mathrm{\Nu}} +\ep^2 \Dd_{\Nu}^{ (\mathrm{per})}$ has finite band and, when left multiplied  with  a column vector, acts as a convolution with a compact kernel. Unlike CNNs, the diffusive nature of these convolution kernels give them  more physical motivation.

It is well known that  the type of activation function has a huge impact in  qualities of the model, its optimization, and   mathematical properties concerning approximation theory \citep[Chapter 8]{DeepLearning}. 
We highlight that the complexity of PSBC's cost function on trainable weight, i.e., its polynomial degree, increases with $\Nt$ and $\Npt$, making high values of either (especially $\Nt$) prohibitive. Such polynomial dependence however differs from - and contrasts with - standard ANN and CNN models, where activation functions introduce strong nonlinearities into the model.

\footnotetext{Derivative in almost everywhere sense.}

\begin{table}[htbp]
\centering
\begin{tabular}[t]{>{\centering\arraybackslash}p{0.21\linewidth}p{0.21\linewidth}p{0.21\linewidth}p{0.25\linewidth}}
\toprule
 \begin{minipage}{\linewidth}
 \centering
 Sigmoid\\
 $\sigma(z) = \frac{1}{1+e^{-z}}$
 \end{minipage}
 & \begin{minipage}{\linewidth}
 \centering
 $\sigma(z) = \tanh(z)$
 \end{minipage}& \begin{minipage}{\linewidth}
 \centering
 ReLU\\
 $\sigma(z) = \max\{0,z\}$
 \end{minipage} & \begin{minipage}{\linewidth}
 \centering
 PSBC, Eq. \eqref{GeneralFwdhierarchicalAsvec:diffusive}\\
 $\sigma(z=(u,w)) =f(u;w)$
 \end{minipage} \\ 
 \midrule
 \begin{minipage}{\linewidth}
 \centering
 $\sigma(\R)$ bounded \\
 $\partial_z\sigma(\R)$ bounded
 \end{minipage}& \begin{minipage}{\linewidth}
 \centering
 $\sigma(\R)$ bounded \\
 $\partial_z\sigma(\R)$ bounded
 \end{minipage} &\begin{minipage}{\linewidth}
 \centering
 $\sigma(\R)$ unbounded \\
 $\partial_z\sigma(\R)$ bounded \footnotemark
 \end{minipage} & \begin{minipage}{\linewidth}
 \centering
 $ f(\R;w)$ unbounded \\
 $\partial_uf(\R;w)$ unbounded
 \end{minipage} \\ 
\bottomrule
\end{tabular}
  \caption{A few types of activation functions and some of their properties; cf.  \citep[Chapters 6 and  9]{DeepLearning}. \label{table:1}}
\end{table}

Several activation functions are described in Table \ref{table:1}, where properties of  their 0th and 1st order derivatives - which differ substantially - are summarized. Indeed, both sigmoid and hyperbolic tangent  activation functions are bounded, and their first order derivatives decay fast towards zero, making the learning process slow when trainable weights are too large; for these reasons they are sometimes called \textit{squashing nonlinearities}. In contrast, the PSBC has polynomial activation functions. Activation functions of ReLU type  differ because their (almost everywhere) derivatives are bounded, which is not necessarily the case for the  PSBC. 
 
Last, it must be said that the architecture of the PSBC bears similarities to (one to one) Recurrent Neural Networks (RNNs), especially in the case of the PSBC with layers-$\Nt$-shared architecture, whose definition is recursive (from last layer down to input layer). Another point of similarity is that, in the PSBC construction, both $\dtp$ and $\dtu$ are fed to the model before forward propagation, like hidden parameters fed to an RNN's initial layer; in such  case, the fixed values $\dtp$ or $\dtu$ would be modified  - or rather ``distorted'' - by  weights after each backpropagation, and then propagate through the RNN network. But this analogy stops here: RNN's weights are optimized with the sole intention of improving the model with respect to a cost function. In addition to that, the feature's growth during forward propagation is tamed with squashing nonlinearities. Differently,  in the PSBC  both $\dtu$ and $\dtp$ are adjusted by the Invariant Region Enforcing Condition \eqref{IREC} to tame features' growth during forward propagation; cf. \citep[Chapter 10, especially 10.7 and 10.8]{DeepLearning}.

\section*{Supplementary Material, Data and Code availability}
 All the code for this paper has been written in Python and is available on Github \citep[]{Bin_phase_github}, which also contains Supplementary Material. Relevant data is available at ``\cite{Bin_phase_data}''.
 
\acks{
The author would like to thank the hospitality of the Indiana University-Bloomington (U.S.A.), where he had several interesting conversations about gradient flows with Peter Sternberg. 
}
%
%
\part*{Appendix}
\appendix
%
%
\section{Initialization, optimization, hyperparameter tuning, and hardware}\label{app:InitializationOptmHyper}
\paragraph{Hardware and computational statistics.} To make replication and verification simpler, small computations were performed using Google Colab notebooks with TPUs. Unavoidably lengthier numerical computations (mostly related with training the model on the MNIST database) were performed  on a supercomputer at MathAM-OIL using 6 cores over a single node. 

\paragraph{MNIST database.}\label{app:mnist} This database consists of $70,000$ images of handwritten digits from $0$ to $9$, each with size $28\times 28$ pixels, stored as matrices of the  same dimension. Following a historical trend in the ML field, the first 60,000 individuals were chosen for train-development purposes, while the last $10,000$ individuals make up the test set. For further details, see  ``\cite{Mnist}''.

\paragraph{Initialization.}\label{app:initialization} Trainable weights  $W_u^{\fl{\cdot}} \in \R^{\Npt}$ in \eqref{fullmodela} and $W_p^{\fl{\cdot}} \in \R^{\Np}$ are initialized  randomly, as Normal variables with mean $0.5$ and diagonal covariance matrix, with variance $\sigma^2$; in all the simulations, we took $\sigma =0.1$.  Parameters $\dtu$ and $\dtp$ values contemplated during grid search were $\{0.1, 0.2\}$; although they have always been  initialized with the same value, they evolve independently, as described in Section \ref{sec:invariant}.

\paragraph{Optimization.}\label{app:opt} We rely  on several different optimization techniques. 

Learning rates obey shrinking schedules, being discounted at a fixed rate every 5 epochs.

Model optimization uses (minibatches) Stochastic Gradient Descent (SGD), with minibatches of size $32$.  We do so due to limited computer memory, and also because adding noise to the parameter search (by subsampling the sample space) improves optimization. Minibatch size was chosen based on empirical studies pointing out that small size minibatches yield better performance \citep[]{masters2018revisiting}. 

Number of epochs was 10 (grid search) and 20 (training), although optimization may stop earlier because \textit{early stopping} was employed \citep[Chapter 7.8]{DeepLearning}. In this way,  weights are recorded at their best value (according to a cost function being  minimized), and optimization halts if the cost function does not achieve better results in the next $p^*$ iterations; the quantity $p^*$ receives the name \textit{patience}. In all the simulations, Accuracy \eqref{accuracyformula} was used as the monitored quantity. Patience was set as  $p^* = 10$ in both grid search and training.
%
%
\section{Some discrete maximum principles}\label{app:MP}
The results in this section are all standard: they are discrete counterparts to classical results in elliptic PDE theory \cite[Chapter 3]{gilbarg2015elliptic}.  We prove  them here for the sake of completion, referring the reader to ``\cite[Theorem 12.5.1]{Strikwerda}'' for a different approach.

In what follows, let $V \in \R^{N}$, for $N\geq 3$. For any $j \in \{1, \ldots, N\}$, define
\begin{equation}\label{Averageperator}
\begin{split}
\average_{j}\left(V\right) := \left\{ \begin{array}{ccc}
      V_2, & if &j =1; \\
      \frac{V_{j-1}+ V_{j+1}}{2}, & if &1< j <N; \\
      V_{N-1}, & if &j =N.
    \end{array}\right.
\end{split}
\end{equation}
We begin by proving an auxiliary result:
\begin{Lemma} \label{Lemma:MP} Let $V \in E = \R^N$. Consider $\Dd_{\mathrm{N}}$ be the diffusion finite difference matrix as in \eqref{DifferenceMatrixNeumann}. Then the following statements are true:
\begin{enumerate}[label=(\roman*), ref=\theTheorem(\roman*)]
\item\label{Lemma:MP:Itema} The condition $\left(\Dd_{\mathrm{N}} V\right)_j\leq 0$ is equivalent to $\average_{j}\left(V\right) \leq V_j$.
\item\label{Lemma:MP:Itemb} The condition $\left(\Dd_{\mathrm{N}} V\right)_j\geq 0$ is equivalent to $\average_{j}\left(V\right) \geq V_j$.
\item \label{Lemma:MP:Itemc} If $\displaystyle{V_m = \min_{1\leq j \leq N} \{ V\}}$, then $\displaystyle{\left(\Dd_{\mathrm{N}}V \right)_m \geq }0$. Likewise, 
if $\displaystyle{V_M = \max_{1\leq j\leq N} \{ V\}}$, then $\left(\Dd_{\mathrm{N}}V \right)_M \leq 0$.
\item \label{Lemma:MP:Iteminvariance} (Invariant positive cone) Given $U \in \R_+^N$ and $\alpha \geq0$ then $\displaystyle{\left(\Id_N - \alpha \Dd_{\mathrm{N}}\right)^{-1} U\in \R_+^N.}$

\item  \label{Lemma:MP:norm} $\displaystyle{\Vert\left(\Id_N - \alpha \Dd_{\mathrm{N}}\right)^{-1}\Vert_{\ell^{\infty}\to \ell^{\infty}}\leq 1.}$
\end{enumerate}
\end{Lemma}
The results (iv) and (v) are important in the study of invariant regions \citep[Lemma 3.2.]{Hoff}: the former is related to  elliptic maximum principles, whereas the latter to parabolic maximum principles. 

\begin{Proof}
The proofs of (i)-(ii) are straightforward, hence omitted. Assertion (iii) is a direct consequence of (i) and (ii). We then prove (iv), arguing by contradiction. Assume the existence of $V\in \R_+^N \neq \bm{0}$ such that 
$$\left(\Id_N - \alpha \Dd_{\mathrm{N}}\right)^{-1} U\not \in \R_+^N.$$
Taking $j = \mathrm{argmin}(U)$, it must hold that $\left\{\left(\Id_N - \alpha \Dd_{\mathrm{N}}\right)^{-1}V\right\}_j = U_j <0$. 

Now, let's rewrite the last equality in the equivalent form $V_j = \left\{\left(\Id_N - \alpha \Dd_{\mathrm{N}}\right)U\right\}_j.$
As $V \in \R_+^N$  the right hand side is non-negative, which implies that 
$ \left( \alpha \Dd_{\mathrm{N}}U\right)_j \leq U_j<0.$
However, (iii) implies that $\left(\Dd_{\mathrm{N}}U\right)_j\geq 0$. This contradiction finishes the proof.

The proof of (v) is also by contradiction. Assume that $\displaystyle{\Vert\left(\Id_N - \alpha \Dd_{\mathrm{N}}\right)^{-1}\Vert_{\ell^{\infty}\to \ell^{\infty}}> 1}$. Hence, there exist two vectors  $V, U \in E$ such that
$\displaystyle{\Vert U\Vert\li=1}$, $\displaystyle{\Vert V\Vert\li>1}$ and 
$\displaystyle{\left(\Id_N - \alpha \Dd_{\mathrm{N}}\right)^{-1}U = V.}$
Without loss of generality, let $M = \mathrm{argmax}(V)$ be so that $V_M = \Vert V\Vert\li >1$. Then, using (i), we get that $(\Dd_{\mathrm{N}}V)_M \leq 0$. On the other hand, we combine  $\alpha \geq 0$ with $\displaystyle{U = \left(\Id_N - \alpha \Dd_{\mathrm{N}}\right) V}$ to get 
$$ U_M = \left(\left(\Id_N - \alpha \Dd_{\mathrm{N}}\right)V\right)_M= V_M - \alpha (\Dd_{\mathrm{N}}V)_M \geq V_M >1,$$
in contradiction with  $\Vert U\Vert\li=1$. This finishes the proof of (v). \end{Proof}
\begin{Remark}[Periodic Boundary Conditions]\label{rmk:Periodicbc} The proofs in this session rely essentially on the Maximum Principle. They extend naturally to the case of periodic boundary conditions, substituting $\Dd_{\mathrm{N}}$ in Lemma \ref{Lemma:MP} by $\Dd_{\mathrm{N}}^{ (\mathrm{per})}$ given in \ref{discussion:Periodicdiffusion_matrix} and modifying  the average operator \ref{Averageperator} to

\begin{equation}\label{Averageperator:periodic}
\begin{split}
 \widetilde{\average_{j}}\left(V\right) := \left\{ \begin{array}{ccc}
       \frac{V_{2}+ V_{N}}{2}, & if &j =1; \\
       \frac{V_{j-1}+ V_{j+1}}{2}, & if &1< j <N; \\
       \frac{V_{N-1}+ V_{1}}{2}, & if &j =N.
      \end{array}\right.
\end{split}
\end{equation}

\end{Remark}
%
%
\section{Proofs and auxiliary results}\label{app:proofs}
Throughout this section we shall use the following notation: for any $p\in \R$ and any set $\mathscr{A}\subset \R$, define
$$\displaystyle{\mathrm{dist}(p, \mathscr{A}) := \inf_{q\in \mathscr{A}}\vert p-q \vert}, \quad \text{ and}\quad \displaystyle{\mathrm{diam}(\mathscr{A}) := \sup_{a,b \in \mathscr{A}}\vert a - b\vert}.$$ 
Recalling the notion of convexity from Section \ref{sec:notation}, whenever a set $\mathscr{A}$ is convex  it is easy to show  that for any two points $p$ and $q$ in it and any $C>0$, we have
\begin{equation}\label{convexity}
 \mathrm{dist}\left(p, \mathscr{A}\right)\leq C, \quad \text{and} \quad \mathrm{dist}\left(q, \mathscr{A}\right)\leq C, \quad \text{implies} \quad \mathrm{dist}\left(\lambda p + (1-\lambda) q, \mathscr{A}\right)\leq C,
\end{equation}
therefore the set  $\widetilde{\mathscr{A}} = \{p \in \R \, |\,\mathrm{dist}\left(p, \mathscr{A}\right)\leq C \}$ is also a convex set.  
\subsection{An abstract invariant region lemma}
Throughout the training process it is important to control the $\ell^{\infty}$ norm of trainable weights. 
Thanks to  \eqref{epochiteration}, we approach this problem by showing global existence\footnote{We keep the term ``global''  even when $\Nt <\infty$, because the result is easily extended to the case $\Nt = +\infty$. Furthermore, because $\mathrm{dist}(p, \mathcal{R}) = \mathrm{dist}(p, \overline{\mathcal{R}})$ in the $\ell^{\infty}(\R)$ topology, the result in Lemma \ref{Lem:AbstrInvRegion} still holds when $\mathscr{S}^{\fl{\Nt}}$ has infinitely many points.}  of the dynamics \eqref{fullmodel}, proving that  $\Vert \U_{\cdot}(X, \alpha^{\fl{\cdot}})\Vert\li$ and $\Vert\P_{\cdot}(\frac{1}{2}, \beta^{\fl{\cdot}})\Vert\li$ can both be controlled by the $\ell^{\infty}$-norm of the trainable weights $\alpha^{\fl{\cdot}}$ and $\beta^{\fl{\cdot}}$, independently of  $\Nt$.  

As trainable weights $\alpha^{\fl{\cdot}}$ and $\beta^{\fl{\cdot}}$ vary throughout optimization, we need to understand how they affect the growth of $\U_{\cdot}(X, \alpha^{\fl{\cdot}})$ and $\P_{\cdot}(\frac{1}{2}, \beta^{\fl{\cdot}})$ as the latter propagate through \eqref{fullmodela} and \eqref{fullmodelb}, respectively. Roughly speaking, we understand the range of values that each of these quantities may assume by dividing them in two parts: one that is the convex hull of the trainable weights $\alpha^{\fl{\cdot}}$ and $\beta^{\fl{\cdot}}$, the other as an  ``spill out'' region, whose existence is a side effect of numerical  discretization. 

We begin with an auxiliary result.
\begin{Lemma}\label{lem:polynomial} The polynomial 
$ p(z) = -z^4 + z^3 + 2z -1$
is non-negative in the range $1\leq z \leq 1 + \frac{1}{\sqrt{3}}$.
\end{Lemma}
\begin{Proof}
Rewrite the polynomial as $p(z)= z -(1-z)^2(z^2 + z + 1)$. Then the result follows from $p(z)\geq z -\frac{(z^2 + z + 1)}{3}=:q(z)\geq q(1) = 0$ in $1\leq z \leq 1 + \frac{1}{\sqrt{3}}$.
\end{Proof}
The next result is fundamental in the rest of this Appendix. 
\begin{Lemma}[Abstract invariant region]\label{Lem:AbstrInvRegion} Let $\mathcal{R} \subset \R$ be a convex set of parameters satisfying  $\{0,1\} \subset \mathcal{R}$. For a fixed $\beta \in \mathcal{R}$, let 
\begin{equation}\label{ODEabstract}
\U_{n+1 } = \U_n + \dt\,\U_n (1 - \U_n)(\U_n - \beta). 
\end{equation}
Then, the following properties hold: 
\begin{enumerate}[label=(\roman*), ref=\theTheorem(\roman*)]
\item \label{Lem:AbstrInvRegion:local} If $\U_{n} \in \mathcal{R}$, then $\mathrm{dist}(\U_{n+1}, \mathcal{R}) \leq \dt\,\mathrm{diam}\left(\mathcal{R}\right)^3.$

\item \label{Lem:AbstrInvRegion:bootstrap} For any $\displaystyle{0 \leq \dt \leq \frac{1}{\sqrt{3}\,\mathrm{diam}(\mathcal{R})^2}}$, the set
$\displaystyle{\mathscr{S} := \left\{X \in \R \,\Big|\, \mathrm{dist}(X, \mathcal{R}) \leq \dt\,\mathrm{diam}\left(\mathcal{R}\right)^3\right\}}$
is convex and positively invariant under \eqref{ODEabstract}.
\end{enumerate}
\end{Lemma}
We remark that when compared to classical global existence results in PDEs, Lemma \ref{Lem:AbstrInvRegion:local} and Lemma \ref{Lem:AbstrInvRegion:bootstrap} are, roughly speaking, equivalent to showing local existence (and associated bounds), and then applying a bootstrap argument, respectively; cf. \citep[Theorems 2.1 and 3.9]{SmollerRauch}. 

\begin{Proof}[of Lemma \ref{Lem:AbstrInvRegion}]
Throughout the proof we write $D = \mathrm{diam}\left(\mathcal{R}\right)$. To begin with, \eqref{ODEabstract} gives
\begin{equation}\label{eq_25}
\vert \U_{n+1} - \U_n\vert = \dt \vert\U_n\vert \cdot \vert 1 - \U_n \vert \cdot \vert \U_n - \beta\vert.
\end{equation}
Because $\{ 0,1, \beta, \U_n\} \subset \mathcal{R}$ we can majorize the right hand side by $\dt \, D^3$. As $\U_n \in \mathcal{R}$, the left hand side is bounded from below by $\mathrm{dist}(\U_{n+1}, \mathcal{R})$, and this completes the proof.

Now we turn to the proof of (ii).  We prove positive invariance using an induction argument on $n$. The result is clearly true for $n=0$, since $\U_0 = X \in \mathscr{S}$ due to the normalization condition \eqref{normalizationcondition}. Thus, assuming that  $\U_n \in \mathscr{S}$ for all $0 \leq k \leq n$,  we must prove that $\U_{n+1}\in \mathscr{S}$. 

A simplification is readily available: it suffices to consider the case $\U_n \in \mathscr{S} \setminus \mathcal{R}$, for (i) contemplates the other case. To begin with,  observe  that $\mathscr{S}$ is convex, which it inherits from the convexity of the set $\mathcal{R}$ allied to convexity of the $\ell^{\infty}$ norm (see discussion below \eqref{convexity}). Thus, from  $0 \in \mathcal{R}\subset \mathscr{S}$ we conclude that $\{- \dt\, D^3, + \dt\, D^3\} \subset \mathscr{S}$. Furthermore, as we also have that $\U_n \in \mathscr{S}$, the result follows if we  show that $\U_{n+1}$ belongs to the interval $[- \dt\, D^3, \U_n]$ when $\U_n \geq 0$, or that $\U_{n+1}$ belongs to the interval $[\U_n, + \dt\, D^3] $ when $\U_n \leq 0.$ We shall prove only the former assertion, the proof of the latter being similar.

Since $\U_n \in \mathscr{S} \setminus \mathcal{R}\subset \R$ is non-negative, we must have $\U_n \geq \max\{1, \beta\}$ and, consequently, that $ \dt\,\U_n (1 - \U_n)(\U_n - \beta) \leq 0$; inspecting \eqref{ODEabstract}, this implies that $\U_{n+1} \leq \U_n$. Therefore, in order to finalize the proof it suffices  to show that $- \dt\, D^3 \leq \U_{n+1}$ or, equivalently, that
\begin{equation}\label{Equivalentlhs}
- \dt\, D^3 \leq \U_n\left(1 + \dt (1 - \U_n)(\U_n - \beta)\right)
\end{equation}
holds, where the equivalence is due to  \eqref{ODEabstract}. Since $\U_n$ and $\U_{n+1}$ have the same sign, it holds that $0 \leq \U_{n+1}\leq \U_n$, therefore both $0$ and $\U_n$ belong to the convex set $\mathscr{S}$. Consequently,  we only need to consider the cases in which both quantities $\U_n$ and $\U_{n+1}$ have different signs. %

It is easy to conclude by triangle inequality that  $\mathrm{dist}(\U_n,p) \leq \mathrm{dist}(\U_n, q) + \mathrm{diam}(\mathcal{R})$ for any $p,q \in \mathcal{R}$. By optimization on $q$ and the induction hypothesis, this  yields
\begin{equation*}
\begin{split}
\mathrm{dist}(\U_n,p) &\leq \mathrm{dist}(\U_n, \mathcal{R}) + \mathrm{diam}(\mathcal{R}) \leq D(1 + \dt\, D^2).
\end{split}
\end{equation*}
Thus, we can bound  $\left(1 + \dt (1 - \U_n)(\U_n - \beta)\right)$  from below by $\left( 1 - \dt(D + \dt\, D^3)^2 \right)$, since $\{1, \beta\} \subset \mathcal{R}.$ 
Consequently, 
\begin{align*}
\min\left \{\, (D + \dt\, D^3)\left( 1 - \dt(D + \dt\, D^3)^2 \right),\, 0\, \right\}\leq \U_n\left(1 + \dt (1 - \U_n)(\U_n - \beta)\right).
\end{align*}
It is evident that $-\dt\, D^3 \leq 0$, therefore it suffices to study the values of $\dt$ for which the inequality
\begin{align*}
- \dt\, D^3\leq (D + \dt\, D^3)\left( 1 - \dt(D + \dt\, D^3)^2 \right), 
\end{align*}
holds, which we claim to hold whenever $\displaystyle{ \dt\, D^2 \leq \frac{1}{\sqrt{3}}}$ is satisfied. Indeed, we shall use a rearrangement of the above inequality as
$$- \dt\, D^2 \leq (1 + \dt\, D^2)\left( 1 - \dt\, D^2(1 + \dt\, D^2)^2 \right).$$
Plugging $Z = 1 + \dt\, D^2$ into the  previous inequality and expanding gives $-Z^4+Z^3 + 2Z -1 \geq 0$, which  holds in the range $\displaystyle{1\leq Z \leq 1 + \frac{1}{\sqrt{3}}}$, thanks to the result of Lemma \ref{lem:polynomial}. Unpacking, it gives the equivalent statement $\displaystyle{ \dt\, D^2 \leq \frac{1}{\sqrt{3}}}$, proving the claim. We have  then established \eqref{Equivalentlhs}, and with this we conclude the proof. \end{Proof}

\begin{Corollary}\label{Cor:AbstrInvRegion}
For any sequence $\left(\alpha^{\fl{k}}\right)_{-1\leq k \leq \Nt-1}$, with $a^{\fl{-1}}=1$, define the sets 
$$\mathcal{R}^{\fl{k}} :=\mathrm{conv}\left(\{0,1\} \cup_{m=-1}^{k}\{ \alpha^{\fl{m}}\}\right),$$ 
for $-1 \leq k \leq \Nt-1$. Let $\displaystyle{\U_{k+1} = \U_k + \dt\,\U_k(1-\U_k)(\U_k - \alpha^{\fl{k}}),}$
with $\U_0 \in [0,1]$, and $0 \leq k \leq \Nt-1$. Then, whenever $\displaystyle{0 \leq \dt \leq \frac{1}{\sqrt{3}\,\mathrm{diam}(\mathcal{R}^{\fl{\Nt-1}})^2}}$, we have
$$\mathrm{dist}(\U_k, \mathcal{R}^{\fl{k-1}}) \leq \dt\,\mathrm{diam}\left(\mathcal{R}^{\fl{k-1}}\right)^3, \quad \text{for all} \quad 0\leq k \leq \Nt.$$
\end{Corollary}

\begin{Remark}[the case of parameter range on a symmetric set]\label{rmk:symmetric_trainable_weights} In the case of $\mathcal{R}_{\mathrm{sym}}$ being a symmetric set about the origin (that is, $\beta \in \mathcal{R}_{\mathrm{sym}}$ if and only if $-\beta \in \mathcal{R}_{\mathrm{sym}}$), 
then it is possible to extend the previous proof to show that, whenever $\displaystyle{0 \leq \dt \leq \frac{1}{\mathrm{diam}(\mathcal{R}_{\mathrm{sym}})^2}}$, the set
$$\mathscr{S} := \left\{X \in \R \,\Big|\, \mathrm{dist}(X, \mathcal{R}_{\mathrm{sym}}) \leq \frac{\dt}{2}\mathrm{diam}\left(\mathcal{R}_{\mathrm{sym}}\right)^3\right\}$$
is convex and positively invariant through \eqref{ODEabstract}. In such a case, Corollary \ref{Cor:AbstrInvRegion} applies to
$\mathcal{R}_{\mathrm{sym}}^{\fl{k}} :=\mathrm{conv}\left(\{0,1, -1\} \cup_{m=-1}^{k}\{ \alpha^{\fl{m}}\} \cup_{m=-1}^{k}\{ -\alpha^{\fl{m}}\}\right).$
\end{Remark}
\subsection{Proof of results in Section \ref{sec:reasoning}}
\begin{Proof}[of Proposition \ref{prop:GlobalODEsystem}] It suffices to prove the 1D case ($\Nu = 1$), since the dynamics is decoupled for each index, and the infimum (resp., supremum) of a quantity $U_m$ over two sets $A\subset B$ is so that $\displaystyle{\inf_{m\in B}U_m \leq \inf_{m\in A}U_m}$ (resp., $\displaystyle{\sup_{m\in B}U_m \geq \sup_{m\in A}U_m}$).

For $\mathcal{R}^{\fl{k}}$ as in Corollary \ref{Cor:AbstrInvRegion}, define 
$$\mathscr{S}^{\fl{k}} := \left\{X \in \R \,\Big|\, \mathrm{dist}(X, \mathcal{R}^{\fl{k}}) \leq \dtu\,\mathrm{diam}\left(\mathcal{R}^{\fl{k}}\right)^3\right\}.$$
The equivalence between the interval $\displaystyle{\left[L^{\fl{k}}- \dtu\, M^{\fl{k}}, R^{\fl{k}} + \dtu \, M^{\fl{k}}\right]}$ and the set $\mathscr{S}^{\fl{k}}$ is clear. Applying Corollary \ref{Cor:AbstrInvRegion} then gives the result. 
\end{Proof}

\begin{Proof}[of Lemma \ref{lemma:monotonicity}]
We argue by contradiction. Assume that $\U_{j}\left(X;\alpha^{\fl{j}}\right) \geq \U_{j}\left(\widetilde{X};\alpha^{\fl{j}}\right)$ holds for all  $j \in \{0, \ldots, n\}$  but at $n+1 \leq \Nt$ we get for the first time
\begin{equation}\label{contradiction-2}
\U_{n+1}\left(X;\alpha^{\fl{n+1}}\right) < \U_{n+1}\left(\widetilde{X};\alpha^{\fl{n+1}}\right).
\end{equation}
From Proposition \ref{prop:GlobalODEsystem}, we immediately obtain the bound $\vert \U_n\vert \leq 1 + \dtu$, for all $n \in \{0, \ldots, \Nt\}.$ Now, as $\alpha^{\fl{n}} \in [0,1]$ and $0\leq \dtu \leq \frac{1}{10}$, a straightforward computation shows that 
\begin{equation}\label{contradiction-derivative}
\max_{ \vert U\vert \leq 2,\, \alpha \in [0,1]} \left\vert D_Uf(U,\alpha^{\fl{n}})\right\vert \leq 8\left(1 + \frac{\dtu}{2}\right)^2\leq 9. 
\end{equation}
Using \eqref{motivation}, a simple application of the Intermediate Value Theorem gives
\begin{align*}
\begin{split}
\U_{n+1}(X, \alpha^{\fl{n+1}}) - \U_{n+1}(\widetilde{X}, \alpha^{\fl{n+1}}) &= \U_{n}(X, \alpha^{\fl{n}}) - \U_{n}(\widetilde{X}, \alpha^{\fl{n}}) \\
& \quad + \dtu (D_U f)\left(\theta, \alpha^{\fl{n}}\right)\left(\U_{n}(X, \alpha^{\fl{n}}) - \U_{n}(\widetilde{X}, \alpha^{\fl{n}}) \right).
\end{split}
\end{align*}
for some $\U_n(\widetilde{X},\alpha^{\fl{n}})\leq \theta \leq \U_n(X,\alpha^{\fl{n}})$. From \eqref{contradiction-2} we conclude that the left hand side of the above equality is strictly negative. However, the bound in \eqref{contradiction-derivative} and the assumption $0\leq \dtu\leq \frac{1}{10}$ imply that the right hand side is non-negative. This contradiction implies that \eqref{contradiction-2} cannot happen, and the result is therefore established.
\end{Proof}

\begin{Proof}[of Proposition \ref{prop:Globalpde}] 
The proof is essentially a reprise of that in Lemma \ref{Lem:AbstrInvRegion}, with some subtle changes due to the presence of the diffusion operator; the main argument invokes  Lemma \ref{Lemma:MP:Iteminvariance}, an idea we learned from  \citep[Theorem 3.3]{Hoff}. 

To begin with, observe that there is no diffusion in the component $\P_k(\frac{1}{2},\beta^{\fl{k}})$, hence the proof of this case is contemplated by Corollary \ref{cor:GlobalODEsystem}. Therefore, in the rest of the proof we are left with estimates for the component $\U_k(X, \alpha^{\fl{k}})$ only. 
For all $-1 \leq n \leq \Nt-1$, define
$$\underline{\gamma}^{\fl{n}} := L_{\alpha}^{\fl{n}}- \dtu M^{\fl{n}}, \quad \overline{\gamma}^{\fl{n}} := R_{\alpha}^{\fl{n}} + \dtu M^{\fl{n}}.$$
We have to show that $\underline{\gamma}^{\fl{n}}\bm{1}\leq \U_{n+1}\leq \overline{\gamma}^{\fl{n}}\bm{1}$  holds for every $0 \leq n +1\leq \Nt$.  First, we rewrite \eqref{fullmodela} in its vectorial form \eqref{discretemodelexplicit} as a two-step model,
\begin{equation}\label{discretemodelexplicit:2steps}
 V^{\fl{n+1}} =  \U_{n} + \dtu\,f(\U_{n};\alpha^{\fl{n}} ), \quad\text{and} \quad\U_{n+1} = \mathrm{L}_{\Nu}^{-1}\cdot V^{\fl{n+1}}.
\end{equation}
From here on we resort to an induction argument on $n$.  The result is clearly true for $n=0$, because of the normalization conditions \eqref{normalizationcondition}. Hence, assume that the result holds for all $0 \leq j \leq n $.

Recall from \eqref{DifferenceMatrixNeumann} that $\mathrm{L}_{\Nu} = \Id_{\Nu} -\ep^2 \, \Dd_{\Nu}$. Direct inspection shows that, for all $\gamma \in \R$, 
\begin{equation}\label{L_inverse}
\mathrm{L}_{\mathrm{N}} \cdot \left(\gamma \bm{1}\right) = \gamma \bm{1}, \quad \text{and}\quad(\mathrm{L}_{\mathrm{N}})^{-1} \cdot \left(\gamma \bm{1}\right) = \gamma \bm{1}
\end{equation}
hold, since $\Dd_{\mathrm{N}}\bm{1} =0$ regardless of the boundary conditions, Neumann or Periodic.  In other words, the vector $\gamma \bm{1}$ is an eigenvector to $(\mathrm{L}_{\Nu})^{-1}$, with associated eigenvalue $1$. 

Now, because of the Invariant Region Enforcing Condition \eqref{IREC} we can apply the Corollary \ref{cor:GlobalODEsystem} to the first equation in \eqref{discretemodelexplicit:2steps}, obtaining $\underline{\gamma}^{\fl{n}}\bm{1}\leq V^{\fl{n+1}}\leq \overline{\gamma}^{\fl{n}}\bm{1}$.
In other words, it holds that
$$ V^{\fl{n+1}} - \underline{\gamma}^{\fl{n}}\bm{1}\geq 0, \quad \text{and} \quad \overline{\gamma}^{\fl{n}}\bm{1} -  V^{\fl{n+1}}\geq 0.$$
Now, a left multiplication by $\mathrm{L}_{\Nu}^{-1}$ preserves these inequalities, thanks to invariance of the positive cone in Lemma \ref{Lemma:MP:Iteminvariance}, yielding
$$ \mathrm{L}_{\Nu}^{-1} \left(V^{\fl{n+1}} - \underline{\gamma}^{\fl{n}}\bm{1}\right)\geq 0, \quad \text{and} \quad \mathrm{L}_{\Nu}^{-1}\left(\overline{\gamma}^{\fl{n}}\bm{1} -  V^{\fl{n+1}}\right)\geq 0.$$
Finally, we rewrite the first term using \eqref{discretemodelexplicit:2steps} and apply \eqref{L_inverse} to the second one, obtaining the desired inequalities. This completes the induction argument, concluding the proof. 
\end{Proof}
\begin{Remark}[Continuum versus discrete]\label{rmk:chueh_conway_smoller_hoff}
It is important to contrast the techniques we use and those of the Invariant region theory, mostly applied in the continuum setting \citep{InvariantRegions}. 
In the sequel,  assume that $\partial_t v = F(v)$ represents a reaction-diffusion PDE, and $v_{n+1} = \widetilde{F}(v_{n})$ its discretization. Roughly speaking, the (continuum) theory looks for  sufficient conditions on a set $\Omega$  and on the behavior of $F\big|_{\partial \Omega}(\cdot)$, in such a way that initial conditions for the Cauchy problem that start on $\Omega$ remain on it for all $t\geq 0$. Notably, none of these conditions directly apply to the discrete problem, but have been adapted in some cases \citep{Hoff}. 

In showing that forward propagation is well-defined our goals are the same as those in the continuum theory: we prove that obstructive conditions prevent solutions to leave a certain bounded set (which is analogous to the aforementioned set $\Omega$). Thus, the results of Propositions \ref{prop:GlobalODEsystem} and \ref{prop:Globalpde}, and that of \citep[Theorem 3.3]{Hoff} are equivalent, although (i) we put emphasis on how the meshgrid $\dt$ must  be changed in order to enforce the existence of an invariant region and (ii), the techniques employed in part of the proofs differ. We highlight that lwe have the advantage of knowing both the range of initial conditions (due to the Normalization Condition  \ref{normalizationcondition})  and the structure of the reaction term; thus, in scope, the result in ``\cite{Hoff}'' applies to a wider range of reaction diffusion models. 
\end{Remark}


\end{document}